\documentclass{article}

\usepackage{PRIMEarxiv}

\usepackage[utf8]{inputenc} 
\usepackage[T1]{fontenc}    
\usepackage{hyperref}       
\usepackage{url}            
\usepackage{booktabs}       
\usepackage{amsfonts}       
\usepackage{nicefrac}       
\usepackage{microtype}      
\usepackage{lipsum}
\usepackage{fancyhdr}       
\usepackage{graphicx}       
\graphicspath{{media/}}     
\usepackage{amsmath}

\DeclareMathOperator*{\argmin}{arg\,min}
\usepackage{float}

\usepackage{amsmath,amssymb}
\usepackage{algorithm}
\usepackage[noend]{algpseudocode}

\algrenewcommand\algorithmicrequire{\textbf{Inputs:}}
\algrenewcommand\algorithmicensure{\textbf{Outputs:}}
\makeatletter
\renewcommand{\ALG@beginalgorithmic}{\small}
\makeatother

\title{GUST: Quantifying Free-Form Geometric Uncertainty of Metamaterials Using Small Data
}

\author{
  Jiahui Zheng, Cole Jahnke, Wei ``Wayne'' Chen \\
  J. Mike Walker ’66 Department of Mechanical Engineering\\
  Texas A\&M University \\
  College Station, TX 77843\\
  \texttt{\{calzheng, ctjahnke03, w.chen\}@tamu.edu} \\
}

\begin{document}
\maketitle

\begin{abstract}
This paper introduces \textit{GUST (\underline{G}enerative \underline{U}ncertainty learning via \underline{S}elf-supervised pretraining and \underline{T}ransfer learning)}, a framework for quantifying free-form geometric uncertainties inherent in the manufacturing of metamaterials. GUST leverages the representational power of deep generative models to learn a high-dimensional conditional distribution of as-fabricated unit cell geometries given nominal designs, thereby enabling uncertainty quantification. To address the scarcity of real-world manufacturing data, GUST employs a two-stage learning process. First, it leverages self-supervised pretraining on a large-scale synthetic dataset to capture the structure variability inherent in metamaterial geometries and an approximated distribution of as-fabricated geometries given nominal designs. Subsequently, GUST employs transfer learning by fine-tuning the pretrained model on limited real-world manufacturing data, allowing it to adapt to specific manufacturing processes and nominal designs. With only 960 unit cells additively manufactured in only two passes, GUST can capture the variability in geometry and effective material properties. In contrast, directly training a generative model on the same amount of real-world data proves insufficient, as demonstrated through both qualitative and quantitative comparisons. This scalable and cost-effective approach significantly reduces data requirements while maintaining the effectiveness in learning complex, real-world geometric uncertainties, offering an affordable method for free-form geometric uncertainty quantification in the manufacturing of metamaterials. The capabilities of GUST hold significant promise for high-precision industries such as aerospace and biomedical engineering, where understanding and mitigating manufacturing uncertainties are critical.
\end{abstract}


\section{Introduction}

Metamaterials are engineered to exhibit specific macroscopic properties, such as exceptional mechanical, thermal, or optical behaviors, through meticulously designed micro- or mesoscale structures (geometry). However, manufacturing processes inevitably introduce geometric imperfections due to factors including limited tool precision, thermal effects, or material variability, leading to uncertainty in the final geometry. These uncertainties, if neglected during design, can critically compromise performance and reliability, especially for safety-critical industries (e.g., aerospace and healthcare) and when material behavior is sensitive to geometric variations.


Accurately quantifying geometric uncertainties in metamaterials is both challenging and essential, due to their typically complex geometries and the high sensitivity of geometry to material properties. Traditional uncertainty quantification (UQ) focuses on \textit{dimensional uncertainty}, limiting analysis to a few key parameters such as lengths, diameters, and angles \cite{6c5c5ea8ed344f1bb0a268b293d6907d, 10.1115/1.4041251, 10.1121/10.0009162}. Despite efforts to model the full geometric uncertainty, they often rely on \textit{predefined uncertainty} models such as uniform boundary dilation/erosion and Gaussian random fields \cite{article, article2, Wang:19, korshunova2021uncertainty}. While computationally tractable, these methods often fail to accurately represent the complex nature of real-world geometric variability, which can be inherently \textit{``free-form''} (e.g., arbitrary boundary deformation and topological changes) and thus beyond predefined uncertainty models.
The fundamental challenge in accurately capturing real-world, free-form geometric uncertainties stems from their high dimensionality. To address this, recent work has leveraged the expressive power of generative adversarial networks (GANs) to learn complex, high-dimensional data distributions. While this deep generative model-based approach shows promise, its effectiveness is often limited by the need for large datasets---an impractical requirement in many contexts where data acquisition necessitates physical fabrication and ex-situ measurement~\cite{10.1115/1.4055898}. 

\begin{figure*}[h]
\centering\includegraphics[width=1\linewidth]{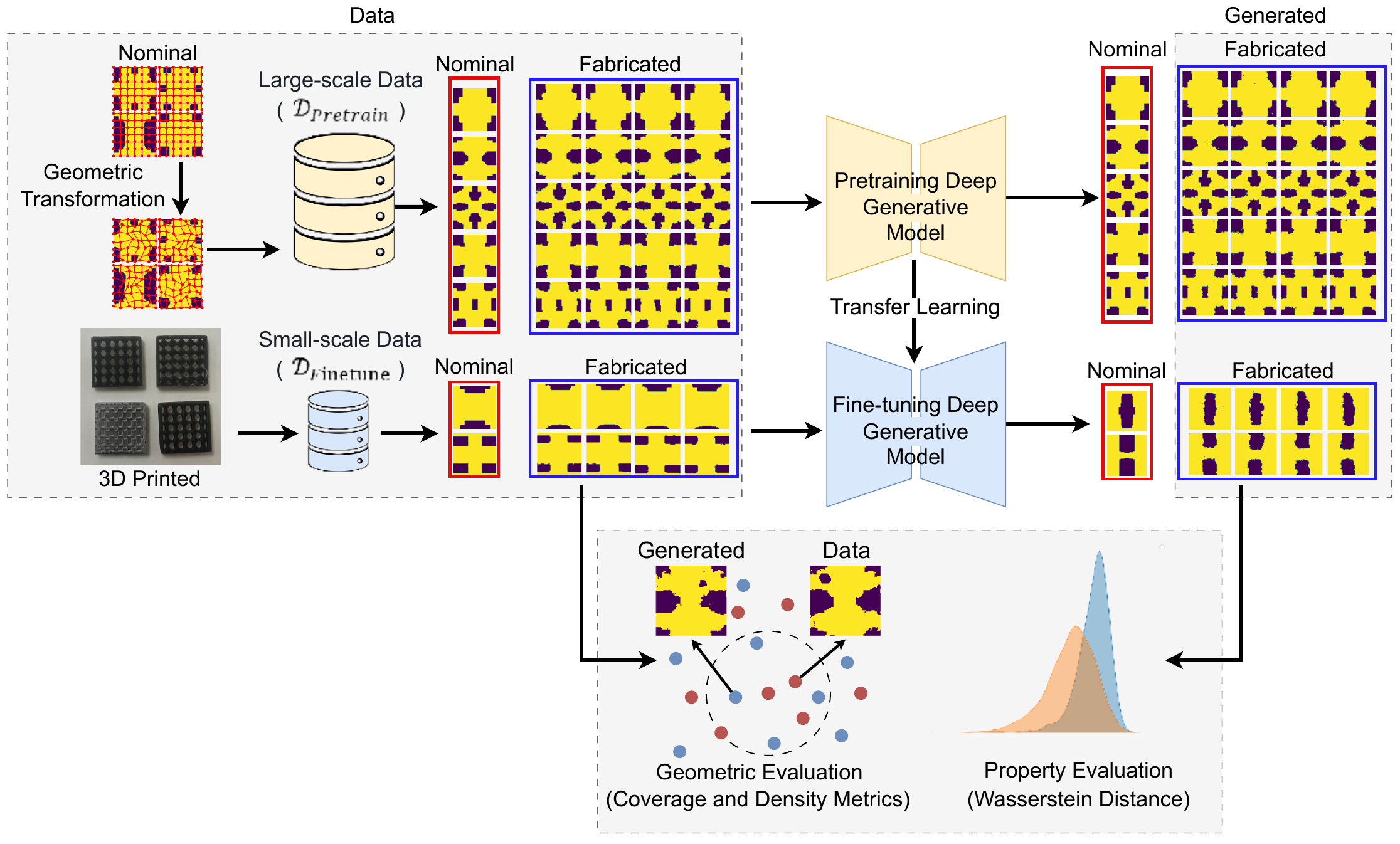}
\caption{Overview of GUST. 
We apply various geometric transformations (e.g., free-form deformation) to add shape and topological variability to nominal designs (i.e., unit cell geometries) and mimic geometric imperfections resulting from manufacturing processes. This enables the construction of a large-scale synthetic dataset \(\mathcal{D}_{\text{pretrain}}\), containing nominal designs and their corresponding synthetic ``as-fabricated'' geometries. The pretraining phase trains a conditional generative model on \(\mathcal{D}_{\text{pretrain}}\) to learn to approximate \(p(\mathbf{x}_{\text{fab}} \mid \mathbf{x}_{\text{nom}})\). The transfer learning phase fine-tunes the pretrained model with a small dataset, \(\mathcal{D}_{\text{finetune}}\), containing real as-fabricated geometries, shifting the generative distribution to the actual real-world geometric uncertainty. Using the resulting generative model to generate Monte Carlo samples of as-fabricated geometries, we can efficiently quantify the geometric uncertainty of fabricated metamaterials.
The performance of the generative model is evaluated in terms of both geometry (using coverage and density metrics) and material properties (using Wasserstein distance).}
\label{fig: workflow}
\end{figure*}

To reduce the data requirements and improve the training stability of generative adversarial networks, we propose \textit{\underline{G}enerative \underline{U}ncertainty learning via \underline{S}elf-supervised pretraining and \underline{T}ransfer learning (GUST)}, with a conditional diffusion model backbone. GUST circumvents the need for extensive real-world manufacturing data by combining self-supervised pretraining and transfer learning. Specifically, as shown in Fig.~\ref{fig: workflow}, we first pretrain the model on large-scale synthetic data, which mimics the as-fabricated geometric deviations resulting from the manufacturing process and can be efficiently constructed at scale from nominal designs. This training approach, akin to \textit{self-supervised learning}, leverages the nominal design data to generate supervisory signals (i.e., corresponding synthetic ``as-fabricated'' geometries), circumventing the need for actual manufacturing data. We then fine-tune the pretrained model with small-scale real-world data. The goal is to predict geometric uncertainty introduced during manufacturing for any given nominal design, i.e., the conditional distribution $p(\mathbf{x}_{\text{fab}}|\mathbf{x}_{\text{nom}})$, where $\mathbf{x}_{\text{fab}}$ is the as-fabricated geometry and $\mathbf{x}_{\text{nom}}$ is the corresponding nominal design. Here, $\mathbf{x}_{\text{fab}}$ is high-dimensional regardless of the dimension of $\mathbf{x}_{\text{nom}}$ due to the free-form nature of realistic uncertainty. By pretraining the deep generative model on a synthetic dataset, the network captures a conditional distribution similar to $p(\mathbf{x}_{\text{fab}}|\mathbf{x}_{\text{nom}})$. 
Fine-tuning the generative model with sparse real-world data will eventually shift the generative distribution toward the actual $p(\mathbf{x}_{\text{fab}}|\mathbf{x}_{\text{nom}})$. Using the resulting generative model to generate Monte Carlo samples of as-fabricated geometries, we can efficiently quantify the geometric uncertainty of fabricated metamaterials. The primary contributions of this work are as follows:
\begin{enumerate}
  \item We construct a conditional diffusion model to effectively capture complex free-form geometric uncertainties given any nominal metamaterial design.
  \item We develop a novel pipeline leveraging artificially generated geometries that mimic as-fabricated structures for self-supervised pretraining.
  \item We utilize transfer learning to adapt the pretrained model to the real-world geometric uncertainties of our manufactured metamaterials, enhancing the practicality of deep generative models for capturing free-form geometric uncertainty.
  \item We conduct a comprehensive comparative study that benchmarks GUST against three baseline approaches with both qualitative and quantitative assessments.
\end{enumerate}


\section{Related Work}

In this section, we first review prior work in manufacturing uncertainty quantification (Sec.~\ref{sec:past_work}), which highlights the limitations of existing approaches. Then we introduce the denoising diffusion probabilistic model (DDPM) (Sec.~\ref{sec:ddpm}), which is the backbone of our framework. Finally, we discuss existing transfer learning methods that address the challenge of data scarcity (Sec.~\ref{sec:transfer_learning}).

\subsection{Past Work on Manufacturing Uncertainty Quantification}

\label{sec:past_work}
To contextualize our approach, early efforts in manufacturing uncertainty quantification primarily focused on \textit{dimensional uncertainties}, employing statistical analyses to address variability in a limited number of design parameters~\cite{10.1115/1.4041251, 10.1121/10.0009162, 6c5c5ea8ed344f1bb0a268b293d6907d}. 
For instance, Nath et al.~\cite{6c5c5ea8ed344f1bb0a268b293d6907d} demonstrate the integration of statistical analyses and Monte Carlo simulations for UQ in fused filament fabrication. They assigned uncertainty to process parameters (e.g., nozzle temperature, layer thickness) and used Monte Carlo sampling to evaluate the impact on the thickness error distribution of the printed part, supporting the selection of process parameters to minimize expected geometric error. Similarly, Morris et al.~\cite{10.1115/1.4041251} modeled manufacturing variations of as-built geometric parameters  (e.g., thickness and height) with multivariate joint probability distributions.
However, dimensional uncertainty becomes less accurate when applied to complex geometries or processes with significant spatial variations, as it may not fully capture nonlinear dependencies or scale-dependent effects in manufacturing variability.


\begin{figure}[h]
\centering\includegraphics[width=.5\linewidth]{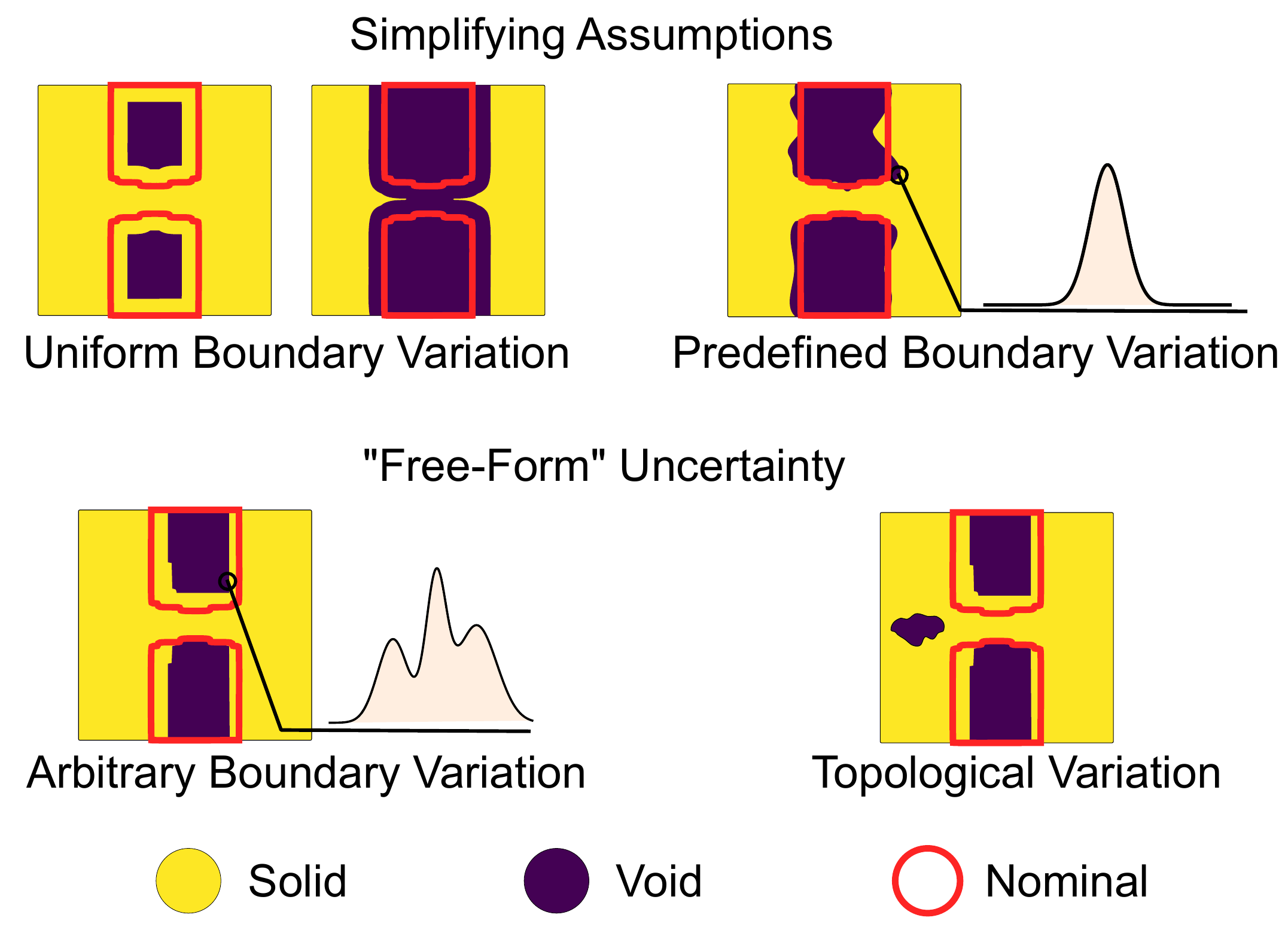}
\caption{Assumptions on the forms of geometric uncertainties. Prior approaches typically depend on predefined simplifying assumptions to model geometric variability, whereas GUST automatically learns arbitrary boundary and topology variations without such simplifying assumptions.}
\label{fig: variation}
\end{figure}

For more complicated geometries or processes, to model the full geometric uncertainty without incurring prohibitive costs, prior work often resorts to simplifying assumptions and relies on \textit{predefined} uncertainty models, which may not fully capture the complexity of real-world variations. For instance, Wang et al.~\cite{Wang:19} modeled fabrication‐induced shape errors as uniform boundary erosion and dilation (Fig.~\ref{fig: variation}). 
Chen and Chen~\cite{article2} modeled geometric uncertainty as a normal velocity perturbation of the level‐set boundary. This velocity perturbation is assumed to be a Gaussian random field along the boundary. Korshunova~\cite{korshunova2021uncertainty} constructed a binary random‐field model whose statistics (e.g., two-point correlations and volume fraction) match the CT scans of as-manufactured lattice struts. Sampling this binary random field yields ensembles of “statistically equivalent” microstructures. These predefined uncertainty models become less accurate when applied to complex free-form shapes and high-dimensional uncertainty spaces, where the increased complexity and variable interactions demand more advanced techniques.

Recent advancements in uncertainty quantification have leveraged machine learning techniques, particularly deep generative models such as Generative Adversarial Networks (GANs)~\cite{NIPS2014_5ca3e9b1}, to model more complex, ``free-form'' geometric uncertainties \cite{10.1115/1.4055898}. Here, free-form means that no assumption is required on the form of geometric uncertainties, thereby allowing both arbitrary boundary variation and topological variation (Fig.~\ref{fig: variation}). Generative models like GANs provide a powerful approach to capture complex data distributions and therefore, enable the learning of high-dimensional, free-form geometric uncertainties.
Despite their capability, deep generative methods typically require extensive and diverse datasets for training. This poses a challenge for many manufacturing scenarios due to the high cost associated with data acquisition.

Built on the idea of using deep generative models to quantify free-form geometric uncertainty, this work combines self-supervised pretraining and transfer learning to reduce data requirements while maintaining the generative model's high capacity of capturing real-world geometric uncertainty.


\begin{figure}[h]
\centering\includegraphics[width=.6\linewidth]{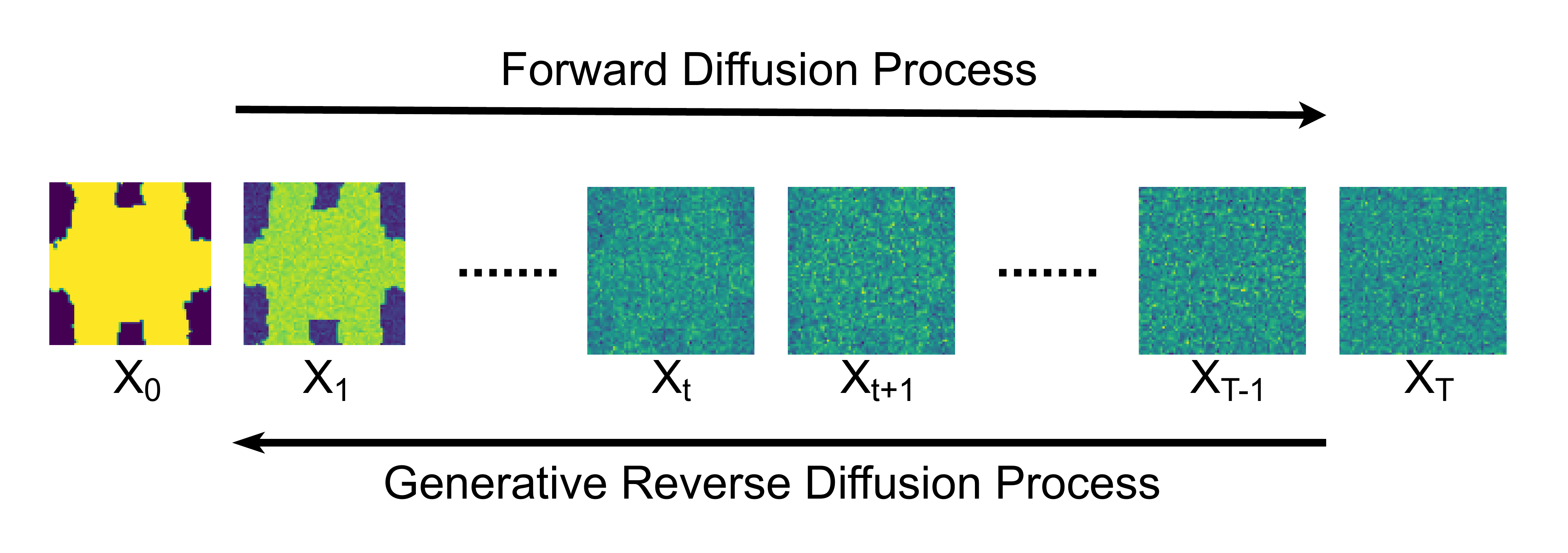}
\caption{The forward diffusion process adds noise to data and the reverse process recovers the original data through iterative denoising. Here, $t=0,...,T$ is the timestep.}
\label{fig:forward}
\end{figure}

\subsection{Denoising Diffusion Probabilistic Models (DDPM)}
\label{sec:ddpm}
Denoising diffusion probabilistic models have emerged as a powerful deep generative modeling approach, often demonstrating superior performance compared to GANs across various tasks~\cite{NEURIPS2020_4c5bcfec,Song2021MaximumLT,karras2022elucidatingdesignspacediffusionbased, NEURIPS2021_49ad23d1,pmlr-v37-sohl-dickstein15}. Diffusion models employ two Markov chains: a forward process that gradually adds noise to data and a reverse process that denoises the noisy data back to the original data distribution (Fig. \ref{fig:forward}). Specifically, given a data distribution \( x_0 \sim q(x_0) \), the forward Markov process generates a sequence of noisy variables \( x_1, x_2, \dots, x_T \) via:
\begin{equation}
q(x_t | x_{t-1}) = \mathcal{N}(x_t; \sqrt{1 - \beta_t}x_{t-1}, \beta_t I), \quad t = 1, \dots, T,
\end{equation}
where \(\beta_t \in (0,1)\) is a hyperparameter controlling noise level. The forward distribution at any timestep \(t\) given the original data \(x_0\) can be written as:
\begin{equation}
q(x_t | x_0) = \mathcal{N}(x_t; \sqrt{\bar{\alpha}_t} x_0, (1 - \bar{\alpha}_t) I),
\end{equation}
where \(\alpha_t = 1 - \beta_t\) and \(\bar{\alpha}_t = \prod_{s=1}^{t}\alpha_s\). Given \(x_0\), we can directly sample \(x_t\) using:
\begin{equation}
x_t = \sqrt{\bar{\alpha}_t} x_0 + \sqrt{1 - \bar{\alpha}_t} \epsilon, \quad \epsilon \sim \mathcal{N}(0, I).
\end{equation}

The reverse denoising Markov chain is defined as a parameterized Gaussian transition:
\begin{equation}
\label{eqn:p_theta}
p_{\theta}(x_{t-1}|x_t) = \mathcal{N}(x_{t-1}; \mu_{\theta}(x_t, t), \Sigma_{\theta}(x_t, t)),
\end{equation}
where \(\mu_{\theta}(x_t, t)\) and \(\Sigma_{\theta}(x_t, t)\) are neural networks trained to minimize the variational lower bound (VLB) of the log-likelihood. The objective function for training the reverse process is simplified to:
\begin{equation}
\label{eqn:L}
L(\theta) = \mathbb{E}_{x_0, \epsilon, t}\left[ \|\epsilon - \epsilon_{\theta}(\sqrt{\bar{\alpha}_t}x_0 + \sqrt{1 - \bar{\alpha}_t}\epsilon, t)\|^2 \right],
\end{equation}
where 
\(\epsilon_{\theta}(x_t, t)\) is the predicted noise from the neural network at timestep \(t\).

In this work, we aim to learn a conditional distribution of as-fabricated geometries given any nominal design. Diffusion models are inherently capable of modeling conditional distributions, which is crucial for guided synthesis processes based on input conditions such as text prompts, semantic maps (e.g., pixel-wise object or scene annotations), class maps (e.g., pixel-wise category labels), or bounding boxes~\cite{9878449,saharia2022paletteimagetoimagediffusionmodels}. Conditioning can be achieved through various methods, including cross-attention mechanisms to incorporate non-spatial conditioning information like text or class labels into the diffusion process, or direct concatenation with the U-Net input for spatial inputs like images, semantic maps, class maps, and inpainting regions (e.g., areas to be filled or reconstructed). 

\subsection{Transfer Learning for Generative Model}
\label{sec:transfer_learning}
Early research on transfer learning in computer vision~\cite{sharif2014cnn,yosinski2014transferable} shows that early convolutional layers learn very general, low-level visual features—edges, corners, simple color or texture patterns—that are broadly useful across almost any image-classification task. Since these filters are not tied to specific object classes, we can safely freeze them and reuse their weights with little risk of overfitting. By contrast, deeper layers build up higher-level, task-specific patterns (object parts or whole-object prototypes) that vary more between datasets, so we typically fine-tune those to adapt the model to new classes or domain. However, uncertainty in our problem could arise in both low- and high-level features. It is unclear at what level of features the difference exists between geometric uncertainties underlying the pretraining and fine-tuning data.

Transfer learning’s success in computer vision inspires its use in generative models, where limited data and overfitting pose similar challenges. To overcome the limitations of training generative models, transfer learning has been increasingly adopted to adapt pretrained models to new tasks. Training a generative model from scratch or fine-tuning a pretrained generative model on limited target domain data often leads to substantial performance decline due to overfitting and memorization~\cite{ouyang2024transferlearningdiffusionmodels}. This is mainly because the training data lacks diversity, failing to cover the full underlying data distribution. To address this, various studies have proposed methods for generative transfer learning. Specifically, transfer learning strategies for generative models can be grouped into two main types: fine-tuning lightweight adapters~\cite{Moon2022FinetuningDM,xie2023difffitunlockingtransferabilitylarge,yang2023one} and fine-tuning with regularization~\cite{zhu2023fewshotimagegenerationdiffusion,ouyang2024transferlearningdiffusionmodels, NEURIPS2022_58ce6a4b,zhao2023closerlookfewshotimage}.

Fine-tuning lightweight adapters involves adjusting a small subset of model parameters, like adapters or specific layers, to efficiently adapt generative models with minimal data, reducing computational cost and overfitting. For instance, Moon et al.~\cite{Moon2022FinetuningDM} proposed a time-aware adaptor embedded within the attention block, constituting roughly 10\% of the diffusion model’s parameters, significantly reducing the number of parameters to adjust. Similarly, Xie et al.~\cite{xie2023difffitunlockingtransferabilitylarge} focused on fine-tuning only specific parameters, such as those tied to bias, class embedding, normalization, and scale factors, for efficient adaptation. Yang et al.~\cite{yang2023one} explored one-shot generative domain adaptation, leveraging lightweight adapters to achieve parameter-efficient model adaptation with minimal data. 
 
Fine-tuning with regularization employs techniques like contrastive or adversarial losses to stabilize model adaptation and prevent overfitting when training on limited data. For example, Zhang et al.~\cite{NEURIPS2022_58ce6a4b} and Zhao et al.~\cite{zhao2023closerlookfewshotimage} focused on one-shot and few-shot generative tasks, respectively, using regularization techniques such as contrastive learning and adversarial losses to ensure stable and generalizable adaptation. Regularizers such as L2-Starting-Point (L2-SP)~\cite{xuhong2018explicit} were proposed to avoid large deviations from the initial pretrained weights and mitigate catastrophic forgetting.

Some studies specifically focus on conditional generative models or diffusion models. In particular, Cheng et al.~\cite{cheng2025provable} showed a theoretical study on how the representation of conditions learned from source tasks can improve the sample efficiency of target tasks in transfer learning.
Ouyang et al.~\cite{NEURIPS2024_f782860c} proposed modeling the score function on the target domain as the score function on the source domain with additional guidance. The guidance network learns the density ratio of the target and source domain data distributions via a binary classifier.
Huijben et al.~\cite{huijben2024denoising} proposed unconditional pre-training to learn general representations, followed by conditional fine-tuning. The conditions (labels) are only incorporated in the fine-tuning stage by adding the trainable label embeddings to time step embeddings.






\section{Methodology}

In this section, we introduce the entire pipeline of GUST, including pretraining dataset construction (Sec. \ref{sec:pretrain_data}), self-supervised pretraining (Sec.~\ref{sec:Self-Supervised}), fine-tuning data acquisition (Sec. \ref{sec:finetunedata}), and transfer-learning (Sec. \ref{sec:transferlearningmethod}). We also introduce how the resulting DDPM can be used for uncertainty quantification in designing metamaterials (Sec. \ref{sec:uqmethod}).

\subsection{Pretraining Dataset Construction}
\label{sec:pretrain_data}
To avoid extensive manufacturing data collection, we stochastically generate synthetic ``as-fabricated'' geometries from as-designed (nominal) ones. Specifically, 
we define a family of stochastic perturbation operators
$
\{\,\mathcal{F}_i \,\}_{i=1}^n$, each represents a specific geometric transformation such as shape deformation, boundary smoothing, dilation, or erosion. Let \( \mathbf{x}_\text{nom} \) be a nominal geometry, a synthetic as-fabricated geometry $\mathbf{x}_\text{fab}$ is obtained by applying a sequence of $l$ operators to $\mathbf{x}_0$, i.e., $\mathbf{x}_\text{fab} = \mathcal{F}_{i_1}\circ \mathcal{F}_{i_2}\circ \cdots \circ \mathcal{F}_{i_l}(\mathbf{x}_\text{nom})$, where $i_1, \ldots, i_l$ are sampled from a pre-defined distribution over $\{ 1, \ldots, n \}$. This allows applying a random sequence of stochastic operators to the nominal design to generate an unlimited number of synthetic as-fabricated geometries. 

In this work, we use two operators that can be expressed in the following forms:
\begin{enumerate}
    \item Free-form deformation (FFD): The geometry is embedded in an \(m\times m\) control‐lattice \(\{P_{ij}\}\). Introduce Gaussian perturbations \(\xi_{ij}\sim\mathcal{N}(\mathbf{0},\sigma^2\mathbf{I})\) to each control point by:
    \begin{equation}
     \mathcal{F}_{\mathrm{FFD}}(\mathbf{x};\sigma,u,v)
     = \sum_{i=0}^{m-1}\sum_{j=0}^{m-1}
       B_i^{m-1}(u)\,B_j^{m-1}(v)\,\bigl(P_{ij} + \xi_{ij}\bigr),
       \label{eqn:FFD}
    \end{equation}
   where \(B_i^{m-1}(t)=\binom{m-1}{i}t^i(1-t)^{m-1-i}\) are Bernstein polynomials and \((u,v)\) are the normalized parametric coordinates of each mesh point.
   \item Random Hole Nucleation: This process randomly samples a seed location \(\mu=(\mu_x,\mu_y)\) uniformly in the interior, and a covariance \(\Sigma\in\mathbb{R}^{2\times2}\), and a scale \(\alpha>0\). To ensure that the covariance matrix \(\Sigma\) is symmetric and positive semi-definite, 
we sample the diagonal entries $\Sigma_{ii} = \xi_i^2, \quad
\Sigma_{jj} = \xi_j^2$, $\xi_i,\;\xi_j \sim \mathcal{N}\bigl(\mu,\;\sigma^2\bigr)$, and constrain 
the off-diagonal term \(\Sigma_{ij}\) such that \(|\Sigma_{ij}| < \sqrt{\Sigma_{ii} \Sigma_{jj}}\), 
guaranteeing valid elliptical contours for the Gaussian-shaped void. Define
\begin{equation}
\mathcal{F}_{\mathrm{hole}}(\mathbf{x};\mu,\Sigma,\alpha,p)
= \mathbf{x}(p) - \alpha\,\exp\!\Bigl(-\tfrac12\,(p-\mu)^\top\Sigma^{-1}(p-\mu)\Bigr),
\quad p\in\mathbb{R}^2.
\label{eqn:hole}
\end{equation}
   Here, $p$ denotes the spatial coordinate at which the signed distance field of the original shape \(\mathbf{x}(\cdot)\) is evaluated. This carves out a smooth, elliptical void in the material.

\end{enumerate}

One can define morphological dilation/erosion and boundary‐blurring operators using standard image processing techniques based on kernel operations and convolution.

By construction, \(\mathcal{D}_{\mathrm{pretrain}}\) can be efficiently generated to an arbitrarily large size, providing paired nominal and synthetic “as‐fabricated” geometries for self‐supervised representation learning.

\subsection{Self-Supervised Pretraining}

\label{sec:Self-Supervised}

After constructing $\mathcal{D}_\text{pretrain}$, we use it to train a conditional deep generative model. This training approach leverages the nominal design data to generate ``as-fabricated'' geometries as supervisory signals and hence resembles self-supervised learning.
Although this self-supervised pretraining process does not learn the actual geometric uncertainty, it captures two pieces of useful knowledge from the synthetic data: (1)~the global structure variability learned from both nominal and ``as-fabricated'' geometries and (2)~a conditional distribution that approximates the actual $p(\mathbf{x}_{\text{fab}} \mid \mathbf{x}_{\text{nom}})$, under certain assumptions of the uncertainty defined by the geometric operations mentioned in Sec.~\ref{sec:pretrain_data}.
One can choose any conditional deep generative model to capture the high-dimensional distribution of $p(\mathbf{x}_{\text{fab}} \mid \mathbf{x}_{\text{nom}})$, where $(\mathbf{x}_{\text{nom}}, \mathbf{x}_{\text{fab}}) \in \mathcal{D}_\text{pretrain}$. When using a DDPM, Eqs. (\ref{eqn:p_theta}) and (\ref{eqn:L}) will become:
\begin{equation}
p_\theta(\mathbf{x}_{\text{fab},{t-1}}\mid \mathbf{x}_{\text{fab},t}, \mathbf{x}_{\text{nom}})
= \mathcal{N}\bigl(\mathbf{x}_{\text{fab},{t-1}};\,\mu_\theta(\mathbf{x}_{\text{fab},t},t, \mathbf{x}_{\text{nom}}),\;\Sigma_\theta(\mathbf{x}_{\text{fab},t},t, \mathbf{x}_{\text{nom}})\bigr)
\end{equation}
and
\begin{align}
L(\theta)
= \mathbb{E}_{\mathbf{x}_{\text{fab},0},\epsilon,t}\Bigl[\;\bigl\|\epsilon \;-\;\epsilon_\theta(\mathbf{x}_{\text{fab},t},t,\mathbf{x}_{\text{nom}})\bigr\|^2\Bigr],
\quad
\end{align}
where $x_t=\sqrt{\bar\alpha_t}\,x_0+\sqrt{1-\bar\alpha_t}\,\epsilon$.
In this work, we employ a conditional DDPM with concatenation embedding and SPADE layers~\cite{park2019semanticimagesynthesisspatiallyadaptive} to learn the conditional distribution \(p(\mathbf{x}_{\text{fab}}\mid \mathbf{x}_{\text{nom}})\). By concatenating the nominal geometry \(\mathbf{x}_{\text{nom}}\) to intermediate feature maps at each diffusion step, the denoising network is guided to steer its reverse diffusion trajectory 
$
p_\theta\bigl(\mathbf{x}_{\text{fab},\,t-1}\mid \mathbf{x}_{\text{fab},\,t}, \mathbf{x}_{\text{nom}}\bigr)$
towards outputs that remain faithful to the condition. This process iteratively denoises random noise, conditioned on \(\mathbf{x}_{\text{nom}}\), to capture the complex geometric variations of the fabricated geometry \(\mathbf{x}_{\text{fab}}\). The pseudocode of self-supervised pretraining is shown in Algorithm~\ref{alg:gust-pretrain} of Appendix~\ref{app:pseudocode}.


\subsection{Acquisition of Fine-Tuning Data}
\label{sec:finetunedata}
To adapt the pretrained conditional generative model to characterize real-world manufacturing uncertainties, one needs to construct a small-scale fine-tuning dataset containing nominal designs and the geometries of corresponding physically fabricated components. Depending on the type and length scale of the fabricated components, high-resolution images of the as-fabricated geometries can be obtained through imaging techniques ranging from standard cameras and 3D scanning to CT scanning and scanning electron microscopy (SEM). Next, we can extract the unit cells, representing the fundamental repeating structure of the metamaterial, from the image utilizing image segmentation tools available in standard image processing software. Subsequently, we need to preprocess the unit cell image by converting it to a binary format (to distinguish material and void) and resizing it to conform to the input resolution of the pretrained model. This process yields a fine-tuning dataset $\mathcal{D}_{\text{finetune}}$, tailored to specific metamaterial types, manufacturing methods, and process settings.

\subsection{Transfer Learning}
\label{sec:transferlearningmethod}
This transfer learning process aims to bridge the gap between synthetic and real-world data. In this stage, we fine-tune the pretrained conditional DDPM on the real-world dataset. The pseudocode of fine-tuning is shown in Algorithm~\ref{alg:gust-finetune} of Appendix~\ref{app:pseudocode}. Knowing the level of features (low-level versus high-level) where the main geometric difference lies between the source and target domains informs the most suitable transfer learning strategy. For example, changing the topology from the source to the target domains may require fine-tuning layers that capture high-level features. For the case where we have no prior knowledge regarding the nature of the target domain, the best approach is to try different transfer learning strategies, such as those introduced in Sec.~\ref{sec:transfer_learning}. In our experiment, we explored several transfer learning strategies, among which no obvious distinctions were observed in their performance. Descriptions of the transfer learning strategies and their quantitative results are shown in Appendix~\ref{app:transfer_learning}. Future work will explore additional transfer learning strategies.









\subsection{Uncertainty Quantification}
\label{sec:uqmethod}
To quantify the geometric uncertainty of fabricated metamaterials, we use the resulting generative model to generate Monte Carlo samples of as-fabricated geometries given the target nominal design as the input condition, i.e., $\mathbf{x}_{\text{fab}}^{(i)} \sim p_\theta(\mathbf{x}_{\text{fab}} \mid \mathbf{x}_{\text{nom}})$. Algorithm~\ref{alg:gust-sampling} in Appendix~\ref{app:pseudocode} shows the pseudocode for sampling the fine-tuned DDPM.

The quantified uncertainties facilitate robust design optimization (RDO), where the nominal design is optimized while considering the presence of manufacturing uncertainties. This optimization problem can be formulated as:
\[
\min_{\mathbf{x}_{\text{nom}}} \left[ -F \left( f(\mathbf{x}_{\text{fab}} \mid \mathbf{x}_{\text{nom}}) \right) \right], \mathbf{x}_{\text{fab}} \sim p_{\theta}(\mathbf{x}_{\text{fab}} \mid \mathbf{x}_{\text{nom}}),
\]
where \( f \) maps the geometry to material properties or performance metrics, and \( F \) is any statistics considered in RDO (e.g., lower confidence bound that considers the worst-case scenario).

\section{Experiment Setting}

In this section, we first describe our data‐preparation pipelines for both pretraining and fine-tuning datasets (Sec. \ref{sec:datapreparation}). Then we present the architecture of our conditional diffusion model (Sec.~\ref{sec:architecture}) and the pretraining and transfer learning settings (Sec. \ref{sec:pretrain_TL}). Finally, we introduce the benchmark baselines (Sec.~
\ref{sec:baselines}) and quantitative metrics (Sec. \ref{sec:metrics}) used to assess geometric uncertainty quantification performance.

\subsection{Data Preparation}
\label{sec:datapreparation}


In this work, we created three datasets: a pretraining dataset $\mathcal{D}_{\text{pretrain}}$ and two fine-tuning datasets $\mathcal{D}_{\text{finetune, synth}}$ and $\mathcal{D}_{\text{finetune, real}}$. Both $\mathcal{D}_{\text{pretrain}}$ and $\mathcal{D}_{\text{finetune, synth}}$ contain synthetically generated ``as-fabricated'' unit cell geometries, whereas $\mathcal{D}_{\text{finetune, real}}$ contains real-world as-fabricated geometries.


We used unit cell geometries selected from the \textit{2D orthotropic metamaterial microstructure} dataset~\cite{meta} as nominal designs, and created the pretraining dataset $\mathcal{D}_{\text{pretrain}}$ by applying FFD (Eq.~\eqref{eqn:FFD}) to the nominal geometries. Specifically, each nominal design was deformed into 20 distinct variants, resulting in a total of 60,000 as-fabricated geometries derived from 3,000 unique nominal designs. {While this small number of as-fabricated geometries can be insufficient for independently learning each conditional distribution $p(\mathbf{x}_\text{fab}|\mathbf{x}_\text{nom})$, the conditional DDPM learns from as-fabricated geometries across 3,000 nominal designs. This allows it to capture the covariance between the as-fabricated geometries of different nominal designs, therefore enabling robust learning through generalization, even with limited data per nominal design.

We added randomness to the as-fabricated geometries by adding Gaussian noise with a standard deviation $\sigma = 6$ to the FFD control points. All geometries were represented as binary pixelated images with a resolution of $64 \times 64$. 

We conducted two experiments using different fine-tuning data, a \textit{synthetic} ``as-fabricated'' geometry dataset $\mathcal{D}_{\text{finetune, synth}}$ and a \textit{real-world} as-fabricated geometry dataset $\mathcal{D}_{\text{finetune, real}}$, respectively. We used the synthetic dataset $\mathcal{D}_{\text{finetune, synth}}$ to simplify method validation. We constructed $\mathcal{D}_{\text{finetune, synth}}$ using FFD with $\sigma = 13$ and introduced random holes of varying locations and shapes using Eq.~\eqref{eqn:hole}. Specifically, we generated 64 synthetic variants of ``as-fabricated'' geometries for each of 15 nominal designs not included in $\mathcal{D}_{\text{pretrain}}$ to simulate real-world use cases (i.e., the nominal designs considered in an application-specific scenario can be different from the ones in the pretraining data). 

The real-world fine-tuning dataset $\mathcal{D}_{\text{finetune, real}}$ contains the geometries of 3D-printed metamaterial unit cells with nominal designs also different from the pretraining data.
As shown in Fig.~\ref{fig: data collection}, to construct \(\mathcal{D}_{\text{finetune, real}}\), we converted the binary pixelated images of nominal designs into vector graphics, imported them into computer-aided design (CAD) software as sketches, and scaled them to a standardized unit cell size as needed. We then extruded each sketch by \(2\)\,mm and patterned them into a block of \(5\times5\) unit cells with an outer \(1.5\)\,mm frame to enhance printing accuracy.
By using a Formlabs 3L stereolithography (SLA) printer with Grey V4 photopolymer resin, we can fabricate about 100 blocks per print. 
In only two passes, a total of 180 sample blocks containing 4,500 unit cells corresponding to 20 nominal designs were printed (Fig. \ref{fig: FabImage}). They were then cleaned in isopropyl alcohol, cured, cleaned again, washed in water, and sanded to remove supports.
To capture as-fabricated geometries, photographs of the printed samples were taken on a light table with a fixed position to ensure stable, level images with clear hole visibility and minimal angle distortions. The images were converted from RGB to grayscale, rotated to align the parts based on their boundaries, and cropped to remove backgrounds. Perspective-induced deformation is shown as skewed and compressed unit cell shapes in the captured images, arising when the camera’s viewing angle distorts the apparent geometry of unit cells. To fix this issue, we retained only the inner 3x3 grid to avoid perspective-induced deformation in outer cells. We then segmented each image into a grid of unit cells. To avoid introducing bias into the dataset, we excluded unit cells whose actual geometries could not be reliably captured by the imaging system (such as those with non-penetrating holes). After this filtering process, 17 designs with 64 unit cells each were retained. 
These extracted images composite $\mathcal{D}_{\text{finetune, real}}$.


\begin{figure}[h]
\centering\includegraphics[width=.6\linewidth]{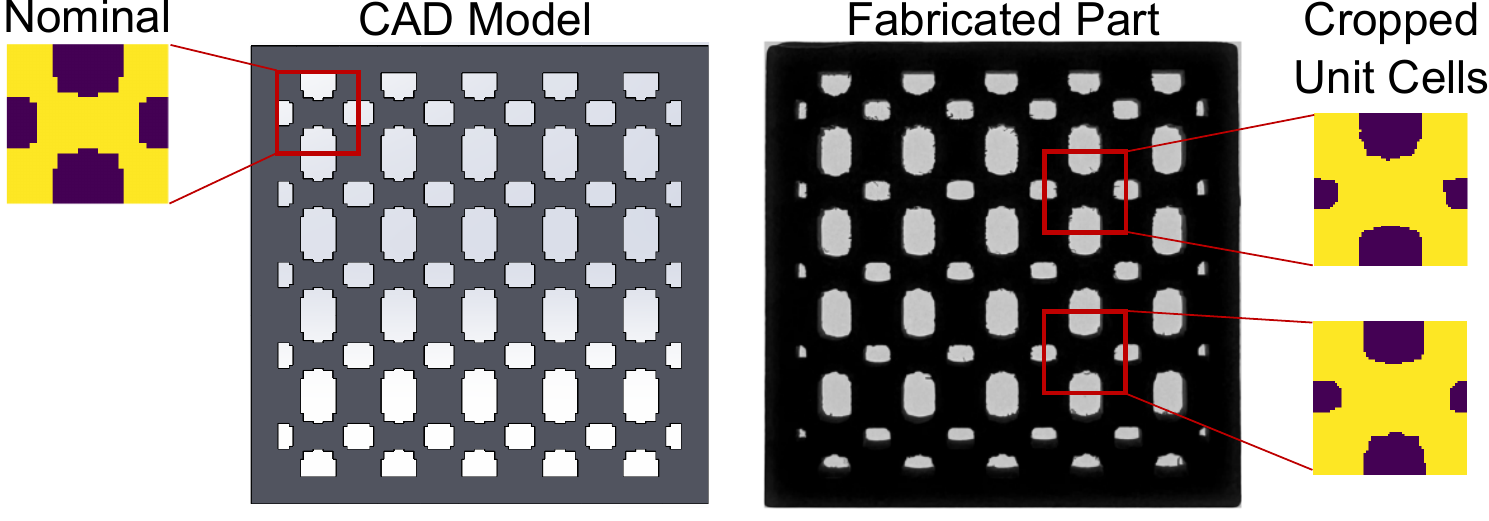}
\caption{Manufacturing data collection process converts a nominal binary array geometry to CAD, extrudes and frames it for 3D printing with SLA resin, followed by cleaning and sanding. Photographs, taken on a light table, are converted to grayscale, rotated, cropped, and segmented into inner unit cells for structural analysis.}\label{fig: data collection}
\end{figure}

\begin{figure*}[h]
\centering\includegraphics[width=1\linewidth]{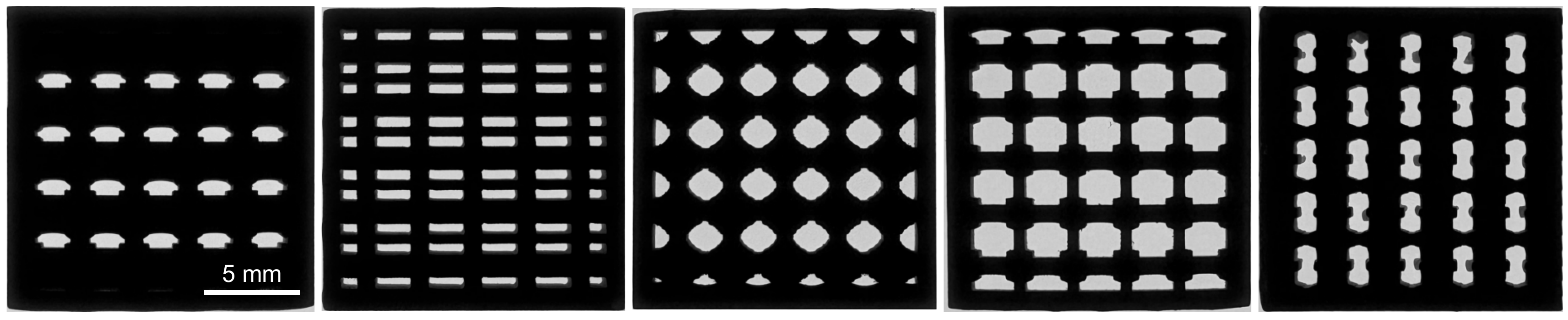}
\caption{Examples of 3D printed metamaterial blocks. A total of 180 blocks were printed in two passes using a Formlabs 3L stereolithography (SLA). }\label{fig: FabImage}
\end{figure*}

\begin{figure*}[h]
\centering\includegraphics[width=1\linewidth]{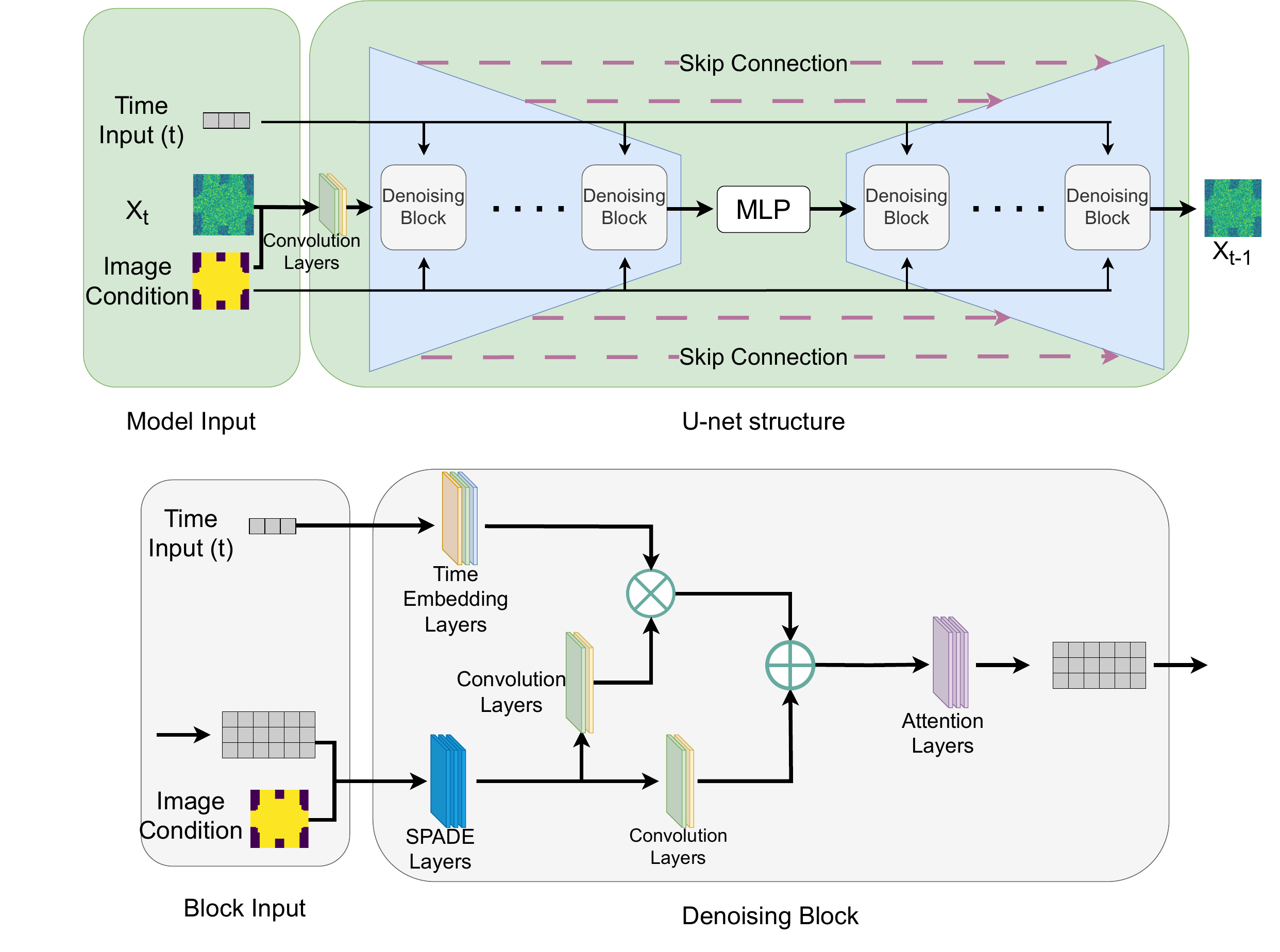}
\caption{The architecture of our conditional diffusion model, consisting of a U-Net structure for conditional generation of as-fabricated geometries from nominal designs. The top diagram illustrates the U-Net framework, featuring a contracting path (encoder) and an expansive path (decoder) with five denoising blocks each, connected by skip connections. A Multi-Layer Perceptron (MLP) at the bottleneck compresses features, while inputs \(\mathbf{x}_t\) (noisy variables) and the image condition (\(\mathbf{x}_{\text{nom}}\)) are processed through convolution layers. The bottom diagram details a denoising block, integrating attention layers for long-range dependencies, convolution layers for spatial features, SPADE layers for conditional normalization, and time embedding layers for temporal diffusion timestep input. The model captures manufacturing uncertainties by conditioning on the nominal geometry and time input, enabling the generation of as-fabricated geometries with stochastic variations.}\label{fig:DDPM}
\end{figure*}

\subsection{Model Architecture}
\label{sec:architecture}
Our conditional diffusion model is implemented using a U-Net-based architecture. This architecture integrates multiple components to learn the reverse diffusion process$p_\theta\bigl(\mathbf{x}_{\text{fab},\,t-1} \mid \mathbf{x}_{\text{fab},\,t}, \mathbf{x}_{\text{nom}}\bigr)$. We detail the key components of the model architecture in the following.

\paragraph*{Contracting Path (Encoder).}
This network receives the noisy sample \(\mathbf{x}_t\) concatenated with the nominal geometry \(\mathbf{x}_{\text{nom}}\) as an “image condition” at the first convolution. It then processes this combined input through five denoising blocks, each progressively reducing spatial resolution (Fig.~\ref{fig:DDPM}). Within each block, attention layers capture long-range dependencies, convolutional layers extract spatial features, SPADE layers apply conditional normalization based on \(\mathbf{x}_{\text{nom}}\)~\cite{park2019semanticimagesynthesisspatiallyadaptive}, and time embedding layers inject timestep information as a “time input.” The encoder thus integrates spatial and temporal conditioning to capture high-level features while downsampling the input.

\paragraph*{Bottleneck.} A multi-layer perceptron (MLP) compresses the data into a low-dimensional representation to extract high-level features from data. It captures the most critical features while maintaining spatial information. 

\paragraph*{Expansive Path (Decoder).} This network contains five denoising blocks that mirror the contracting path, upsampling the data. Skip connections link corresponding encoder and decoder blocks, directly transferring features from the encoder to the decoder to preserve spatial information and enable accurate reconstruction of fine-grained details~\cite{Si_2024_CVPR}.



\subsection{Pretraining and Transfer Learning}
\label{sec:pretrain_TL}

We pretrained the conditional DDPM on $\mathcal{D}_{\text{pretrain}}$ for 180,000 iterations with a batch size of 64. Adam optimizer~\cite{kingma2017adammethodstochasticoptimization} was used with the mean absolute error as the training loss. During training, we set the initial learning rate to be $0.0008$ and decay by $10\%$ after every 5,000 iterations until it reaches 0.000001.  

We then fine-tuned the pretrained model on $\mathcal{D}_{\text{finetune}}$. We experimented with multiple configurations in this study by transferring weights and biases from different layers of the pretrained model (i.e., freezing these layers' parameters during fine-tuning). We fine-tuned each model for 38,400 iterations, the learning rate will decay by 10\% every 960 iterations with the same batch size as in pretraining. Two distinct approaches were investigated as follows.

\paragraph*{Transfer of Specific Denoising Blocks.}
Within the five symmetric denoising blocks in each of the up-sampling and down-sampling pathways of the U-Net, we froze a symmetric subset of these denoising blocks (e.g., freezing the first two up-sampling and the last two down-sampling blocks). Various settings based on this strategy were evaluated; however, the performance differences across these configurations were negligible.

\paragraph*{Transfer of Specific Layers within Blocks.}
Two scenarios were explored in this approach: (1) transferring both the SPADE layers and attention layers across all denoising blocks, and (2) transferring only the attention layers across all blocks. Similar to the previous approach, the differences between performances in these scenarios were not significant.

The fact that different transfer learning strategies yielded comparable outcomes in our case may suggest that the model's performance is relatively insensitive to the specific transfer learning approaches we employed. Our future work will further study the mechanism of transferring different parts of the network and their impact on the generative performance.

All diffusion models were trained on an NVIDIA A100 GPU. Pre-training requires approximately 10 hours, with an additional 2.5 hours for fine-tuning in our experiments. 

\subsection{Baselines}
\label{sec:baselines}
To assess the performance of the transfer learning framework, we evaluate the geometric uncertainty quantification of GUST against three baselines.

\paragraph*{Direct Training (DT) Baseline.} This baseline involves training a diffusion model from scratch on the target dataset without leveraging pretraining and transfer learning with duplicated training configuration from the GUST. In this approach, the model starts with randomly initialized parameters and learns directly from the fine-tuning data. We use the same training configuration as mentioned in Sec.~\ref{sec:pretrain_TL}.

\paragraph*{Dilation-Erosion Baseline.} The dilation-erosion approach~\cite{Wang:19} (Fig. \ref{fig: variation}, top left) approximates manufacturing uncertainty by complementary morphological operations that respectively expand and contract shape boundaries using a predefined structuring element. The expansion (dilation) and contraction (erosion) operators $\mathcal{F}_\alpha$ and $\mathcal{F}_\beta$ are realized by convolving $\mathbf{x}_\text{nom}$ with a square kernel of sizes $\alpha$ and $\beta$, respectively. During the convolution, we compute the maximal (minimal) pixel value overlapped by the kernel and replace the image pixel in the anchor point position with that maximal (minimal) value for dilation (erosion). The parameters $\alpha$ and $\beta$ control how much a geometry dilates or erodes, respectively. In this work, we implement $\mathcal{F}_\alpha$ and $\mathcal{F}_\beta$ using OpenCV's \texttt{dilate} and \texttt{erode} methods. One can also use the approach detailed in Ref.~\cite{Wang:19} to implement the two operators. To ensure a fair comparison, we treat the $\alpha$ and $\beta$ in the dilation and erosion operators as tunable parameters and choose them by using maximum likelihood estimation (MLE), so that the operators' effects accurately reflect the characteristics inherent in the data. Specifically, given a set of nominal designs \(X=\{\mathbf{x}_{\text{nom},i}\}_{i=1}^N\) and a reference collection of real-world fabricated image \(\{\mathbf{x}_{\text{fab},j}\}_{j=1}^M\), we define a Gaussian kernel density estimator:
\[
\hat p_h(\mathbf{x})
\;=\;
\frac{1}{M}
\sum_{j=1}^M
\frac{1}{(2\pi h^2)^{d/2}}
\exp\!\Bigl(-\frac{\|\mathbf{x} - \mathbf{x}_{\text{fab},j}\|^2}{2h^2}\Bigr),
\]
where \(h\) is the kernel bandwidth.
Given a positive integer dilation scale \(\alpha\), we apply \(\mathcal{F}_\alpha\) to each nominal design $\mathbf{x}_{\text{nom},i}$. The log‐likelihood can be computed as
\[
\mathcal{L}_d(\alpha)
=
\sum_{i=1}^N
\log \hat p_h\!\bigl(\mathcal{F}_\alpha(\mathbf{x}_{\text{nom},i})\bigr).
\]
We then obtain the maximum likelihood estimate \(\hat\alpha = \argmin_{\alpha\in[0,100]}\,\mathcal{-L}_d(\alpha)\) by minimizing the negative log‐likelihood via differential evolution~\cite{storn1997differential}. Similarly, the log-likelihood corresponding to the erosion can be expressed as
\[
\mathcal{L}_e(\beta)
=
\sum_{i=1}^N
\log \hat p_h\!\bigl(\mathcal{F}_\beta(\mathbf{x}_{\text{nom},i})\bigr),
\]
which yields the maximum likelihood estimate
\(\hat\beta = \argmin_{\beta\in[0,100]}\,\mathcal{-L}_e(\beta)\).  This procedure produces the dilation and erosion kernel sizes that make the processed geometries closest to real-world distributions, serving as a principled baseline for comparison. 


\paragraph*{Gaussian Random Field (GRF) Baseline.} This approach models geometric variability by adding a GRF to the signed distance function (SDF) of the nominal design~\cite{article}. Let \(\{\mathbf{x}_i\}_{i=1}^N\) be points on a uniform grid within \([0,1]^2\), and define the precision matrix
\[
L^{-1}
= \mathrm{diag}\!\bigl(1/\ell_1^2,\;1/\ell_2^2\bigr).
\]
The covariance matrix \(C\in\mathbb{R}^{N\times N}\) has entries
\[
C_{ij}
= \sigma^2
\exp\!\Bigl(-\tfrac12(\mathbf{x}_i-\mathbf{x}_j)^\top L^{-1}\,(\mathbf{x}_i-\mathbf{x}_j)\Bigr).
\]
We perform the eigen-decomposition \(C\phi_k=\lambda_k\phi_k\), sort the eigenvalues \(\{\lambda_k\}\) in descending order, and truncate the decomposition to retain only the top \(M\) modes.
A single GRF realization is generated via the Karhunen–Loève (KL) expansion,
\[
g(\mathbf{x}_i)
= \sum_{k=1}^M \sqrt{\lambda_k}\,\xi_k\,\phi_k(\mathbf{x}_i),
\quad
\xi_k\sim\mathcal{N}(0,1).
\]

While there are non-Gaussian models such as the field of Gaussian scale mixtures (FoGSM)~\cite{lyu2006statistical}, testing on these models is beyond the scope of this work. In particular, we can place model assumptions along a spectrum. The dilation-erosion and GRF assumptions are at one end of the spectrum, while the completely ``free-form'' geometric variation learned by a deep generative model is at the other end. The non-Gaussian model that can learn from data is in the middle of the spectrum---it has higher flexibility than the strong dilation-erosion and GRF assumptions, yet is not completely ``free-form'' and therefore still risks making incorrect assumptions compared to a deep generative model.
On the other hand, we need the model prediction to generalize to different nominal designs, i.e., given an unseen nominal design, the model needs to predict the corresponding geometric uncertainty without the data of as-fabricated geometries. However, a non-Gaussian model typically does not allow such generalizability without more advanced methodology development.

\subsection{Quantitative Metrics}
\label{sec:metrics}
We use quantitative metrics to evaluate the performance of GUST and the baselines. These metrics were calculated using a test dataset, whose nominal designs were not contained in either the pretraining dataset or the two fine-tuning datasets.

\paragraph*{Geometric Evaluation.}
Precision and recall were earlier attempts to separately measure fidelity (how realistic generated samples are) and diversity (how well they cover the real data's variability)~\cite{NEURIPS2019_0234c510}, but their issues with outlier sensitivity and mode dropping led to the development of the more robust density and coverage metrics. The density and coverage metrics \cite{naeem2020reliablefidelitydiversitymetrics} are designed to evaluate the fidelity and diversity of generated samples by addressing limitations in precision and recall metrics. \textit{Density} quantifies how closely generated samples $\{\,Y\}_{i=1}^M$ cluster around real samples $\{\,X\}_{i=1}^N$, rewarding those in dense regions of the real distribution. It is calculated as the average number of generated samples per real-sample neighborhood spheres $B$ (defined by the k-nearest neighbor distance, $\mathrm{NND}_k(X_i)$) that contain each generated sample $Y_j$, expressed as 
\[
\text{Density} := \frac{1}{k M} \sum_{j=1}^M \sum_{i=1}^N 1_{Y_j \in B(X_i, \mathrm{NND}_k(X_i))}.
\]

\textit{Coverage} measures the fraction of real samples whose neighborhoods contain at least one generated sample, assessing how well the generated samples span the real distribution, and is defined as
\[
\text{Coverage} := \frac{1}{N} \sum_{i=1}^N 1_{\exists j \leq M, Y_j \in B(X_i, \mathrm{NND}_k(X_i))}.
\]
To mitigate the computational expense of computing the density and coverage metrics in the original high-dimensional pixel space, we employed t-distributed stochastic neighbor embedding (t-SNE) with a perplexity of 10. This perplexity was chosen as it led to the most reasonable representation of the local and global structures of real and generated data. This projection into a three-dimensional embedding space preserves local neighbor relationships, a crucial requirement for the accurate calculation of the density and coverage metrics.
    
\paragraph*{Property Evaluation.} 
The ultimate purpose of quantifying geometric uncertainty of metamaterials is usually to quantify their material property uncertainties.
We employ homogenization~\cite{xia2015design} to compute the effective properties, represented by the elastic tensor, of metamaterials under geometric uncertainties.
By computing the Wasserstein distances~\cite{villani2009optimal} between the distributions of each elastic tensor component in generated and real as-fabricated metamaterials, we can directly examine manufacturing uncertainties as reflected in these one-dimensional property distributions.


\section{Results and Discussion}

This section discusses the performance of GUST by comparing its results against baseline approaches in both synthetic (Sec.~\ref{sec:synthexperiment}) and real-world (Sec.~\ref{sec:realworldexperiment}) experiments, focusing on the distribution of generated metamaterial geometries and their homogenized elastic properties. None of the nominal designs used for evaluation was contained in the pretraining or fine-tuning datasets. Through qualitative (visual comparisons of both generated geometries and property distributions) and quantitative analyses (density, coverage, and Wasserstein distance), we demonstrate GUST's ability to capture real-world manufacturing uncertainties and the necessity of transfer learning. 


\subsection{Synthetic Experiment}
\label{sec:synthexperiment}

This experiment used $\mathcal{D}_{\text{finetune, synth}}$ as the fine-tuning data for quick and comprehensive validation of our method.

A qualitative comparison of generated geometries was performed to assess model performance. As shown in Fig.~\ref{fig: Gen}, the ``ground truth'', which are as-fabricated geometries sampled from $\mathcal{D}_{\text{finetune, synth}}$, exhibits noticeable boundary variations and random holes when compared to the nominal design. In contrast, the GRF approach preserves a significant number of sharp corners and straight edges from the nominal design. In certain generated instances (e.g., in the 2nd and 4th rows), an excessive number of irregularly shaped and sized holes appear, deviating from the ground truth. Geometries sampled from both DT and GRF contain artifacts such as noisy and disconnected regions. In addition, the DT baseline also demonstrates challenges in effectively learning the random hole patterns present in the ground truth. On the other hand, the GUST most effectively captures the geometric variability in the ground truth, as evidenced by its boundary deformation pattern and the random holes. 

\begin{figure*}[h]
\centering\includegraphics[width=1\linewidth]{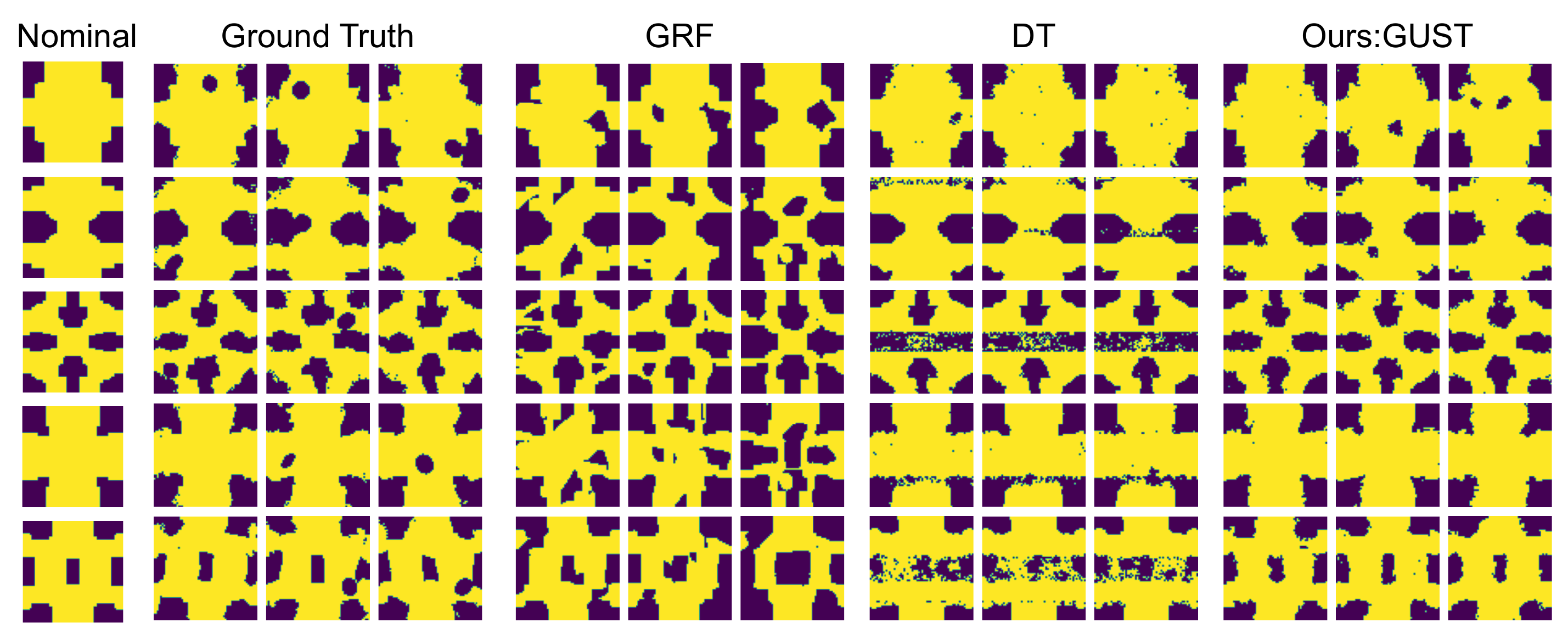}
\caption{
  Comparison of as-fabricated geometries generated by GRF, direct training (DT), and GUST in the synthetic experiment. Yellow and dark regions denote materials and void, respectively. GRF and DT exhibit artifacts like noise and disconnected regions, while GUST more accurately captures real-world uncertainties, such as boundary variation and random holes. 
}
\label{fig: Gen}
\end{figure*}

\begin{figure*}[h]
\centering\includegraphics[width=1\linewidth]{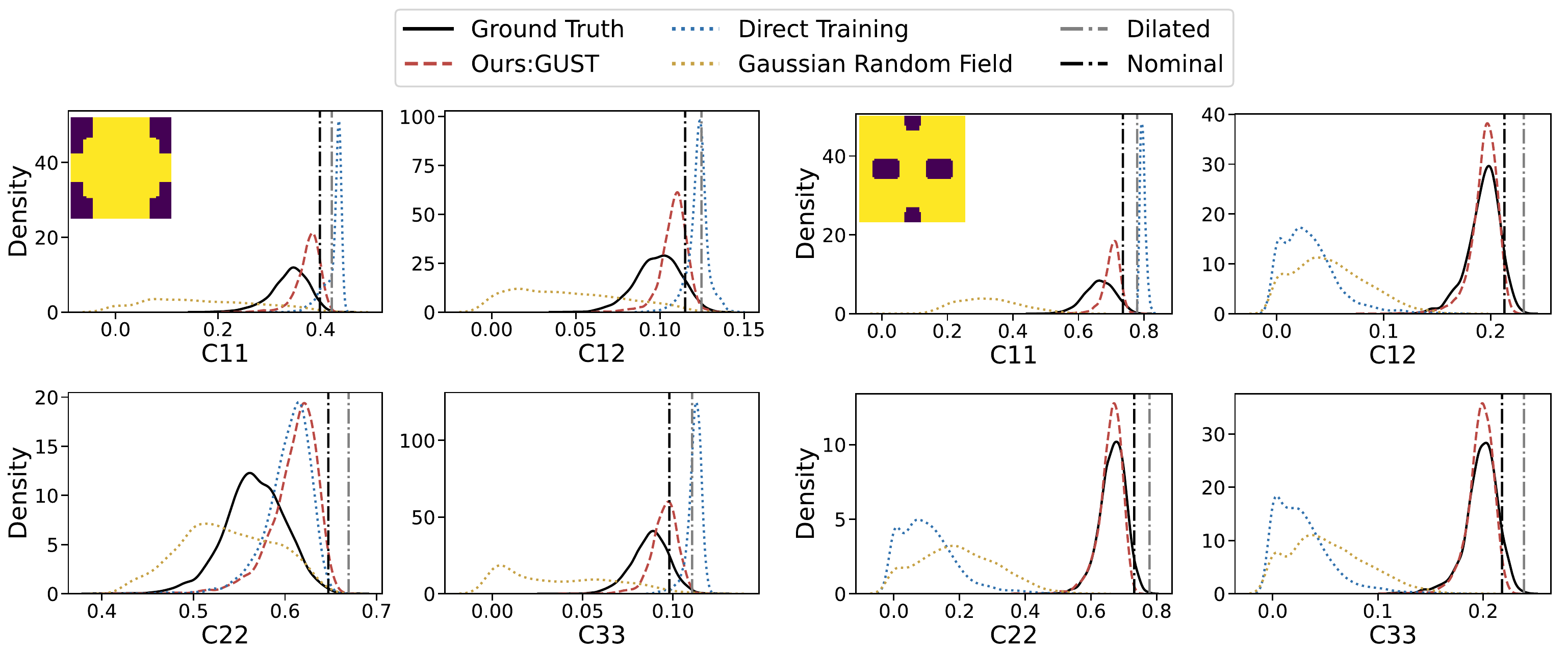}
\caption{Kernel density estimates (KDEs) of homogenized elastic tensor components ($C_{11}$, $C_{12}$, $C_{22}$, $C_{33}$) for two unit cell designs in the synthetic experiment. The corresponding nominal design is shown at the top left corner of each panel. GUST consistently aligns closest to the ground truth across most components.
}
\label{fig: KDE1}
\end{figure*}

\begin{figure*}[h]
\centering\includegraphics[width=1\linewidth]{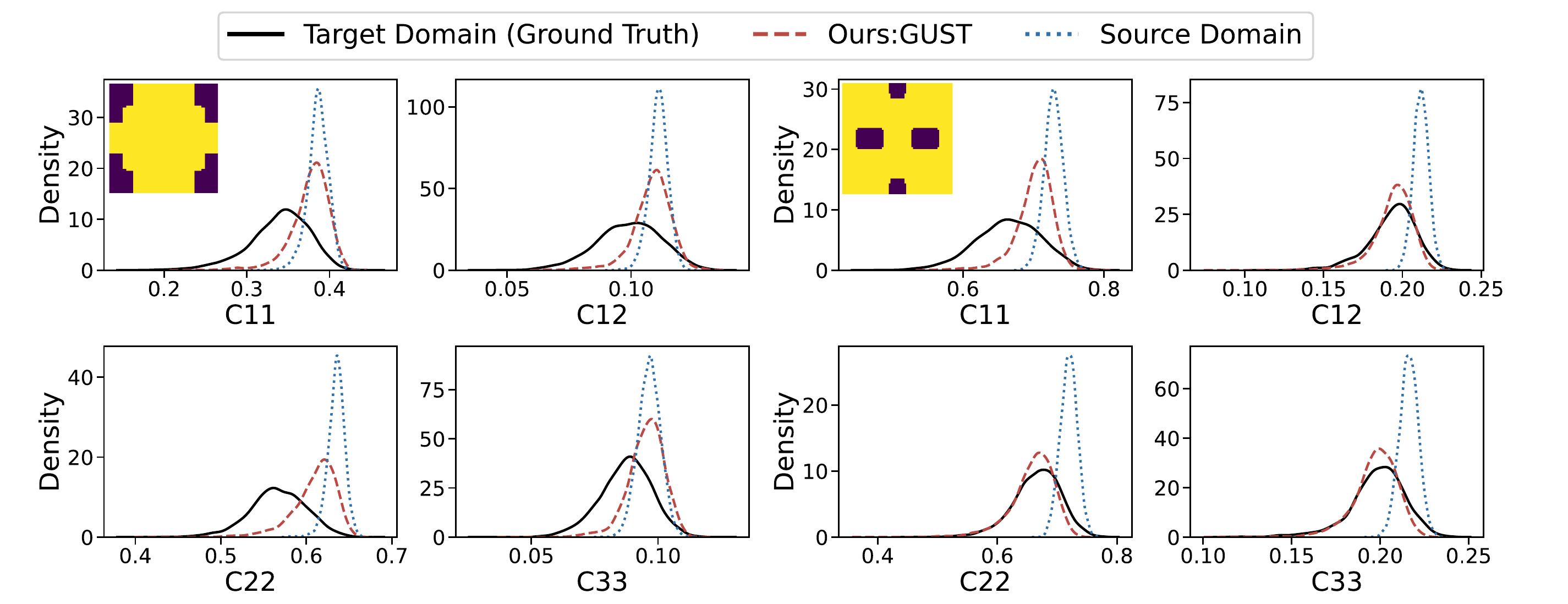}
\caption{
KDEs of homogenized elastic properties for the synthetic experiment showing the discrepancy between the target domain, the source domain, and the distribution learned by GUST. GUST produces a property distribution either closely matching the ground truth or situated between the distributions associated with the source domain and the ground truth.
}
\label{fig: PreKDE1}
\end{figure*}

In addition to generated geometries, we analyzed the elastic properties of geometries generated by GUST and baseline approaches. Figure~\ref{fig: KDE1} presents the distributions of homogenized elastic tensor components ($C_{11}$, $C_{12}$, $C_{22}$, $C_{33}$) corresponding to two nominal designs, where the ``ground truth'' is based on effective properties computed from the geometries in $\mathcal{D}_{\text{finetune, synth}}$. While we picked these representative results, more results are available in Appendix~\ref{appendix:a}. Notably, the dilation-erosion baseline can only lead to point estimates instead of a full distribution. We employed MLE to determine both the dilation and erosion scales, as introduced in Sec.~\ref{sec:baselines}. The resulting dilation and erosion scales are 2 and 1, respectively. Note that a scale of 1 means no change to the nominal geometry. 
Despite optimizing the dilation scale using MLE, it consistently overestimates the actual elastic components (Figs.~\ref{fig: KDE1} and \ref{fig: appendix1}), possibly due to the ignorance of topological changes (i.e., random holes). The GRF approach typically exhibits a markedly different distribution compared to the ground truth, with some exceptions shown in Appendix~\ref{appendix:a}. Given the inherently random nature of GRF, its occasional close alignment with ground truth is expected. 
Compared to GUST, the distributions corresponding to DT are generally farther from the ground truth. DT's performance can be extremely unstable (e.g., Design 2 in Fig.~\ref{fig: KDE1}), possibly due to training instability with limited data.
Overall, the GUST shows a relatively stable performance, evident by its consistently close alignment with ground truth across most of the components and designs. 
These qualitative observations will be further supported by quantitative results via density, coverage, and Wasserstein distance assessment.


To validate that the performance of our GUST framework is not solely due to the similarity between source and target domains, we visualized the property distributions of geometries sampled from the source domain (i.e., domain of pretraining data), the target domain (i.e., ground truth), and generated by GUST (Fig.~\ref{fig: PreKDE1}). Here, as-fabricated geometries from the source domain and ground truth are generated via the same data generation processes as $\mathcal{D}_{\text{pretrain}}$ and $\mathcal{D}_{\text{finetune, synth}}$, respectively. As illustrated in Fig.~\ref{fig: PreKDE1}, despite the difference between the distributions associated with the source domain and the ground truth, GUST produces property distributions that either closely match the ground truth or are situated between the source domain and the ground truth. However, GUST distributions can still be close to those in the source domain in some cases, indicating opportunities for further improvements in future work. The divergence between the source domain and ground truth distributions may influence the performance of the GUST framework. In general, when the source and target domains are closer, it is easier for transfer learning to effectively utilize the features learned from the pretraining model to capture the target distribution. Therefore, to enhance GUST’s performance, it may be beneficial to leverage prior knowledge in constructing the pretraining dataset so that it approximates the target geometric uncertainty as closely as possible.

\begin{figure}[h]
\centering\includegraphics[width=.6\linewidth]{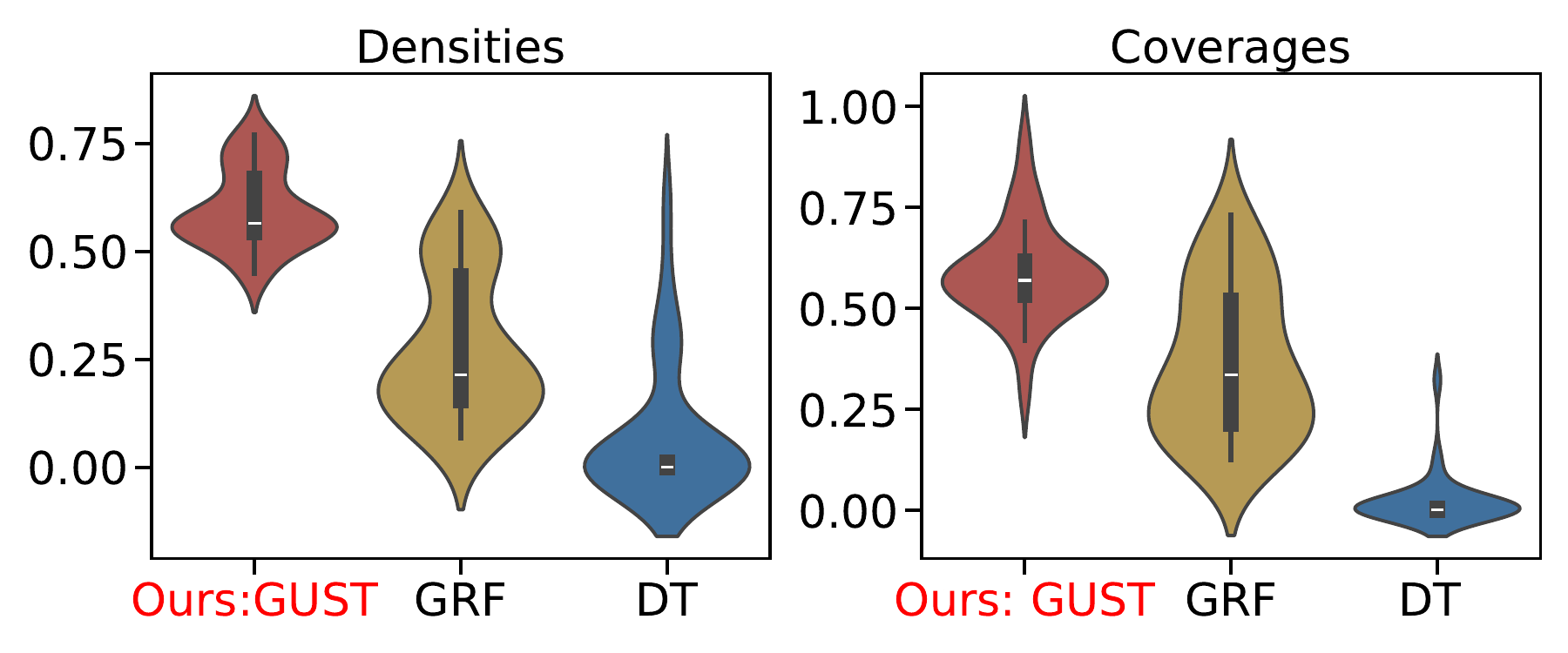}
\caption{Density and coverage (higher is better) metrics for geometries generated by different methods in the synthetic experiment, calculated across 30 nominal designs (each with 3,000 samples of as-fabricated geometries). Density measures sample clustering around real data, while coverage assesses the span of generated samples. 
}
\label{fig: dc-violin}
\end{figure}

We conducted a quantitative analysis of the geometric quality of generated samples through the density and coverage metrics (Sec.~\ref{sec:metrics}). Figure~\ref{fig: dc-violin} shows the distribution of these metrics evaluated on 30 nominal designs, each with 3,000 samples of as-fabricated geometries, ensuring a thorough assessment of the models' performance across diverse scenarios. The density metric quantifies how closely the generated samples cluster around the real data points, while the coverage metric measures how broadly the generated samples span the real data distribution. GUST achieved substantially higher mean density with lower variance than GRF and DT. The p-values in Table~\ref{tab:grf_dt_metrics} indicate a significant difference in the means of the density metric between GUST and the other two baselines. Meanwhile, GUST's coverage metric differs only marginally from GRF (mean values differ by 0.07). 
Based on Fig~\ref{fig: dc-violin}, both GUST and GRF achieve a moderate coverage, meaning that neither of them matches the ground truth well in terms of the distribution of geometries in the reduced space. Considering GUST's generally higher density metrics, we infer that while its high-density regions may overlap with the ground truth, it tends to produce a narrower distribution that lacks coverage. In contrast, GRF receives median scores on both metrics, indicating a possible shift in its distribution's mode compared to the ground truth. A similar coverage to GUST is possibly attributed to GRF's broader distribution. This observation is consistent with the property distributions in Fig~\ref{fig: KDE1}. 


\begin{figure}[h]
\centering\includegraphics[width=.6\linewidth]{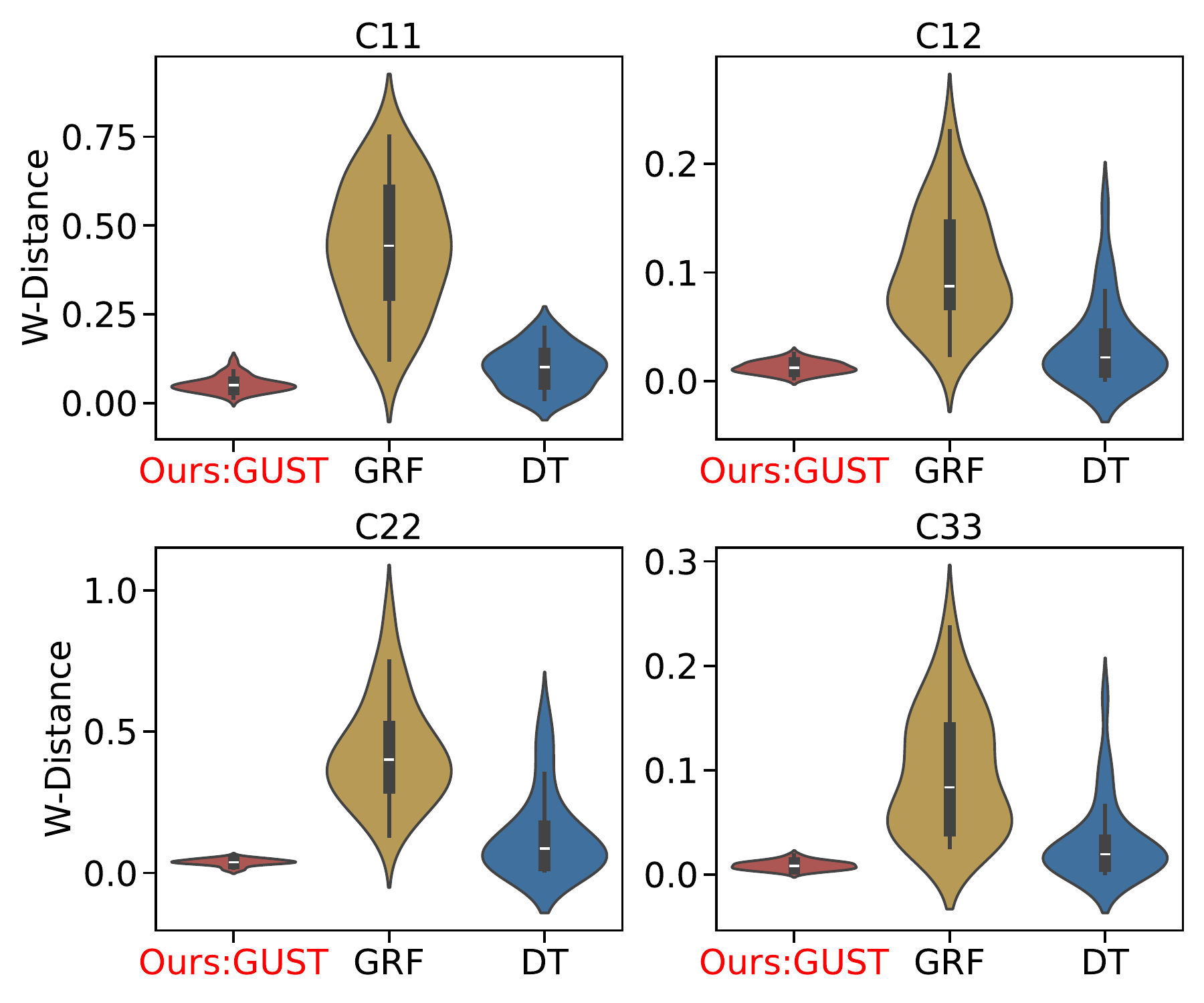}
\caption{Wasserstein distances (lower is better) between property distributions from different approaches and the ground truth distribution of elastic tensor components ($C_{11}$, $C_{12}$, $C_{22}$, $C_{33}$), calculated across 30 nominal designs (each with 3,000 samples of as-fabricated geometries) in the synthetic experiment. GUST consistently shows lower average Wasserstein distances and significantly lower variance.
}
\label{fig: w-violin}
\end{figure}

\begin{table}[ht]
\centering
\caption{P-values comparing the geometry and property distributions represented by GRF and DT against those by GUST.}
\begin{tabular}{lcc}
\toprule
Component & GRF & DT \\
\midrule
Density  & $1.41\times10^{-7}$ & $4.52\times10^{-15}$ \\
Coverage & $2.69\times10^{-1}$ & $8.74\times10^{-10}$ \\
C11      & $1.25\times10^{-3}$ & $2.50\times10^{-3}$ \\
C12      & $1.25\times10^{-3}$ & $2.09\times10^{-2}$ \\
C22      & $2.80\times10^{-3}$ & $6.22\times10^{-3}$ \\
C33      & $1.59\times10^{-3}$ & $1.34\times10^{-2}$ \\
\bottomrule
\end{tabular}
\label{tab:grf_dt_metrics}
\end{table}

We use the Wasserstein distance to measure the alignment between the property distributions of geometries generated by different approaches and the ground truth. Wasserstein distances were calculated using the same generated geometries as those used to calculate the density and coverage metrics (i.e., 30 nominal designs, each with 3,000 generated as-fabricated geometries). Figure~\ref{fig: w-violin} shows the distributions of calculated Wasserstein distances for the elastic tensor components ($C_{11}$, $C_{12}$, $C_{22}$, $C_{33}$). GUST consistently achieves lower mean values with significantly lower variances. The p-values in Table~\ref{tab:grf_dt_metrics} further demonstrate the significant difference in the mean Wasserstein distances between GUST and the two baselines.

\subsection{Real World Experiment}
\label{sec:realworldexperiment}


\begin{figure*}[h]
\centering\includegraphics[width=1\linewidth]{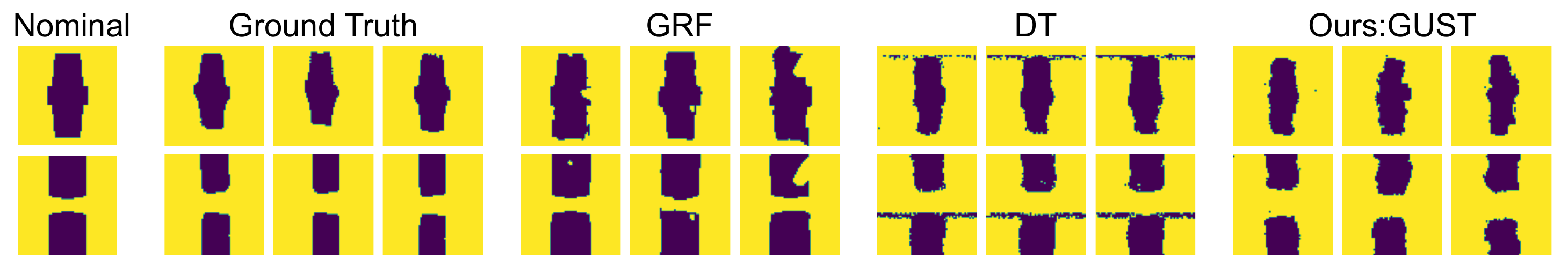}
\caption{
  Comparison of as-fabricated geometries generated by GUST and baselines in the real-world experiment.
  GUST captures geometric variabilities, such as deformed and dilated shapes, that more closely resemble the ground truth compared to GRF and DT.
}
\label{fig: RealGen}
\end{figure*}

\begin{figure*}[h]
\centering\includegraphics[width=1\linewidth]{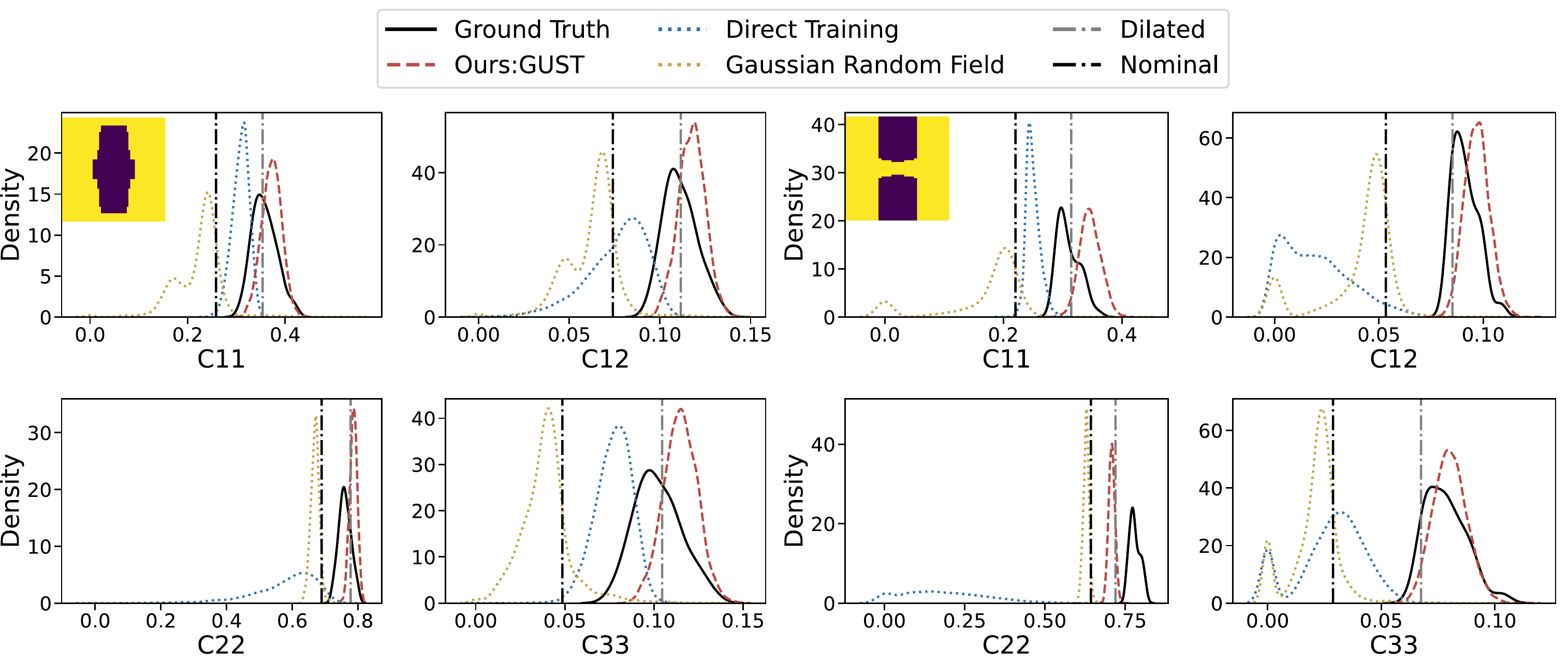}
\caption{
KDEs of homogenized elastic tensor components ($C_{11}$, $C_{12}$, $C_{22}$, $C_{33}$) for two unit cell designs in the real-world experiment. The corresponding nominal design is shown at the top left corner of each panel. GUST produced distributions most closely approximating the ground truth compared to the three baselines.
}
\label{fig: RealKDE1}
\end{figure*}

\begin{figure}[h]
\centering\includegraphics[width=.35\linewidth]{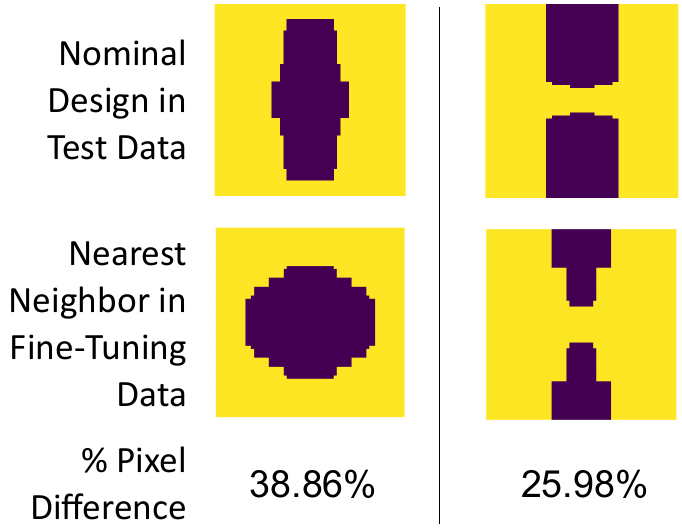}
\caption{Nominal designs in the test data and their nearest neighbors in the fine-tuning data.}
\label{fig: RealGeoComp}
\end{figure}

\begin{figure*}[h]
\centering\includegraphics[width=\linewidth]{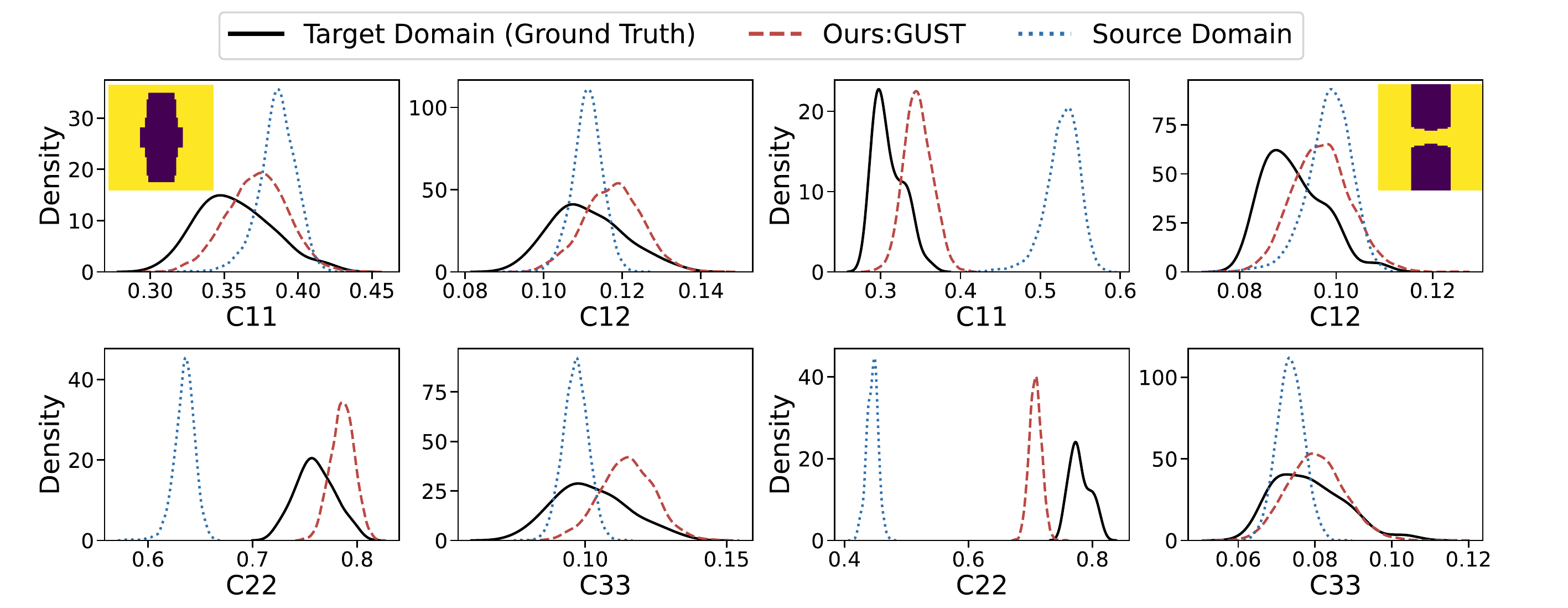}
\caption{KDEs of homogenized elastic properties for the real-world test cases showing the discrepancy between the target domain, the source domain, and the distribution learned by GUST. 
}
\label{fig: RealPreKDE}
\end{figure*}

A qualitative comparison of generated geometries was conducted for the real-world case to assess the performance of geometric uncertainty quantification in actual manufacturing settings. 
As shown in Fig.~\ref{fig: RealGen}, the ground truth shows an obvious dilation shape compared with the nominal. Consistent with results in the synthetic case, the DT and GRF models produce geometries with artifacts, such as noise and disconnected regions. While the GRF generates geometries visually similar to those of GUST, the latter more accurately captures real-world uncertainties, as demonstrated by the dilated shapes in GUST’s geometries closely resembling those in the ground truth. In contrast, the GRF geometries do not show dilated patterns, but with unrealistic abrupt boundary changes. This comparison underscores GUST's superior capability of capturing real-world geometric uncertainties compared to existing approaches.


We computed the homogenized elastic properties of the generated geometries in this real-world experiment and show the distributions of $C_{11}$, $C_{12}$, $C_{22}$, and $C_{33}$ in Fig.~\ref{fig: RealKDE1}. The ``ground truth'' represents the property distribution of data from real-world additively manufactured metamaterials. 
The maximum likelihood estimates for the dilation and erosion scales are 5 and 1, respectively, which indicates that the real as-fabricated geometries predominantly exhibit a dilated pattern. In most cases, the homogenized properties of dilated geometries fall within the high-density region of the ground-truth distributions. This is different from the result in the synthetic experiment, where additional topological changes were considered and were unable to be captured by dilation/erosion. This indicates the limitation of only assuming dilation-erosion as the form of uncertainty---its alignment with the ground truth varies across different scenarios, and it only yields point estimates.
Besides that, the results in Fig.~\ref{fig: RealKDE1} show a generally consistent pattern that matches the synthetic experiment---the property distribution produced by GUST most closely approximates the ground truth compared to the three baselines.

Different from the synthetic experiment, quantifying the density and coverage in this real-world experiment is challenging.
Unlike the perfectly aligned synthetic data, real-world samples introduce unavoidable misalignments from factors like camera positioning and handling. The coverage and density metrics in this research are based on a k-nearest neighbors approach, which is built on distance metrics that are highly sensitive to geometric misalignments. Even a single-pixel shift can cause numerically significant differences between otherwise identical shapes. Given this sensitivity and the limited number of real-world fabricated samples available for validation, achieving high-fidelity results with these metrics is challenging. Our future work will develop new robust methods of evaluating the generative distributions for as-fabricated geometries.

To ensure the discrepancy between the two test cases and the pretraining/fine-tuning data is non-trivial (i.e., the model can generalize to unseen scenarios), we analyze this discrepancy based on two types of differences: 1)~the geometric difference in nominal designs from the fine-tuning and the test data, and 2)~the distributional difference between synthetic and real-world as-fabricated samples.

The first type of difference reflects the generalization capability of the fine-tuned model to unseen nominal designs. Fig.~\ref{fig: RealGeoComp} shows the two nominal designs in the test data and their nearest neighbors in the fine-tuning data. The nearest neighbor is determined by the percentage of different pixels. The relatively large difference (>25\% pixel difference) indicates the level of difficulty in generalizing the fine-tuned model to the two test cases.

The second type of difference represents the deviation of the real-world data distribution (target domain) from the synthetic data distribution (source domain). Due to the difficulty of directly comparing as-fabricated geometry distributions, we visualize the distributions of homogenized elastic properties of as-fabricated geometries (similar to Fig.~\ref{fig: PreKDE1} for the synthetic experiment). The synthetic ``as-fabricated'' geometries of the two test nominal designs were generated using the same geometric perturbation process we used to generate pretraining data (Sec.~\ref{sec:datapreparation}). Figure~\ref{fig: RealPreKDE} shows a noticeable discrepancy between the source and the target domains, while GUST learned distributions much closer to the target domain compared to the source domain.


The results from both experiments underscore the effectiveness of the GUST approach in capturing the geometric uncertainty. By combining pretraining with fine-tuning, both the geometry and material property distributions resulting from GUST consistently showed better alignment with the ground truth distributions compared to the three baseline methods. 




\section{Conclusion}
This work demonstrates that GUST, a framework that combines conditional diffusion models with self-supervised pretraining and transfer learning, can effectively quantify free-form geometric uncertainties with constrained real-world manufacturing data. By comparing with three baseline methods (i.e., dilation-erosion, GRF, and direct training), we demonstrated that GUST can more accurately capture the distributions of as-fabricated geometries, indicated by both qualitative (visual comparisons of generated geometries and property distributions) and quantitative (Wasserstein distance, density, and coverage metrics) assessments. These findings indicate that prior works relying on simplifying assumptions may lead to misleading uncertainty quantification results, and GUST offers a scalable solution for more accurate quantification of real-world uncertainty under data constraints.

Although GUST achieves a closer distributional alignment to the ground truth than existing baselines, a noticeable discrepancy remains in some cases (Figs.~\ref{fig: KDE1} and \ref{fig: RealKDE1}). 
Some recent work has studied the scaling laws that predict the performance from the amount of data and the transfer gap (i.e., the fine-tuning efficiency in the limit of infinite pretraining data). These scaling laws can empirically reveal the convergence of transfer learning performance in general. For example, Mikami et al.~\cite{mikami2021scaling} and Barnett~\cite{barnett2024empirical} derived the scaling law that describes the generalization error using the sizes of pretraining and fine-tuning data, as well as the transfer gap. Hernandez et al.~\cite{hernandez2021scaling} quantified the effectiveness of transfer learning by calculating the effective data ``transferred'' from pretraining (i.e., how much data a model of the same complexity would have required to achieve the same loss when training from scratch), and found the relation between the effective data transferred and the size of fine-tuning data.
Although not derived from GUST, we posit that these scaling laws reflect a general principle: transfer learning performance scales positively with the volume of pre-training and fine-tuning data and negatively with the transfer gap. Accordingly, our future work will expand the size and diversity of our pre-training data to enhance the quality of generated samples.

While we focus on the manufacturing uncertainty of metamaterials, the proposed framework can be generalized to other scenarios, such as quantifying the geometric uncertainty of aerodynamic components or microfluidic devices under wear or erosion. In addition, GUST can be readily extended to 3D or vectorized representations simply by swapping in the appropriate data format and backbone diffusion model. For example, one could employ point‑cloud diffusion networks for unstructured point sets, 3D‑UNet-based diffusion on voxel grids, or mesh-aware diffusion processes on surface meshes. Crucially, these are purely architectural changes---the core GUST recipe of synthetic pretraining, fine‑tuning, and conditional sampling for uncertainty quantification remains unchanged, making the framework fully modular. Combined with advanced imaging techniques such as 3D scanning and micro-computed tomography (micro-CT), GUST can be broadly applicable across diverse use cases.

\section*{Acknowledgments}
Portions of this research were conducted with the advanced computing resources provided by Texas A\&M High Performance Research Computing.


\nocite{*} 

\bibliographystyle{unsrt}   

\bibliography{references} 


\appendix
\section[More]{Additional Results on Property Distributions}\label{appendix:a}
In this section, we provide more results on property distribution comparisons, as shown in Fig.~\ref{fig: appendix1}. Model performance varies across different nominal designs. However, GUST outperforms other baseline methods in most cases.

\begin{figure*}[h]
\centering\includegraphics[width=1\linewidth]{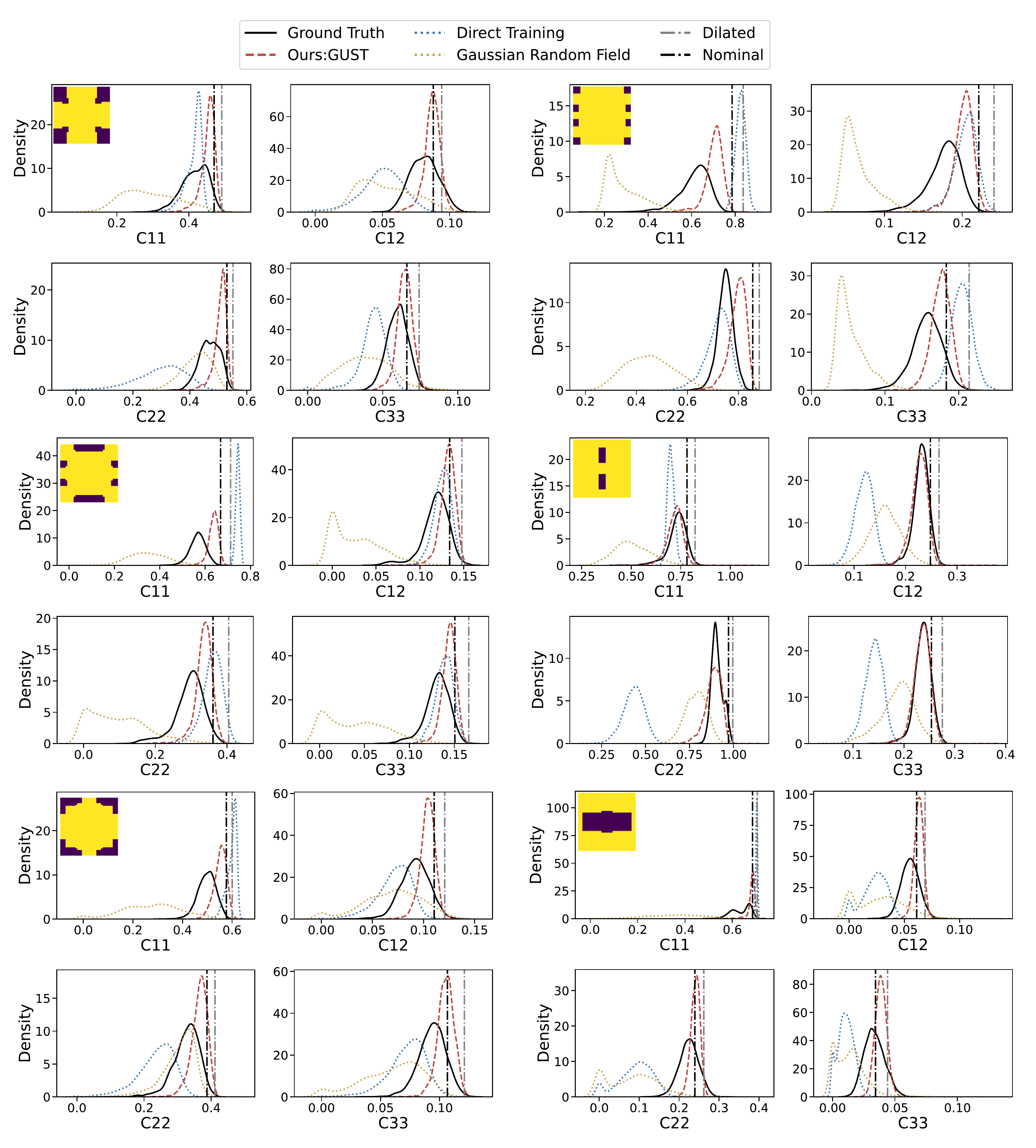}
\caption{KDEs of homogenized elastic tensor components ($C_{11}, C_{12}, C_{22}, C_{33}$) for another 6 designs. The corresponding nominal design is shown at the top left corner of each panel.} \label{fig: appendix1}  
\end{figure*}

\section[More]{Additional Results on Different Transfer Learning Strategies}
\label{app:transfer_learning}

We evaluated five configurations to investigate the impact of transferring different parts of the U-Net. In each configuration, we symmetrically froze the residual blocks and MLP layers in the U-Net and fine-tuned the rest of the network. We summarize the results in Table~\ref{tab:density_coverage}. Each metric's mean and standard deviation were calculated based on 30 nominal designs. 

Across these settings, we do not observe a systematic difference in the statistics of the density and coverage metrics.
This suggests that the difference in geometric variability between pretraining and fine-tuning data was not solely captured in a certain part of the U-Net. This is different from the classical transfer learning settings, where low-level features can be transferable across predictive tasks. Our further study will extend the transfer learning strategies to those mentioned in Sec.~\ref{sec:transfer_learning} and parameter-efficient fine-tuning (PEFT) techniques such as LoRA~\cite{hu2022lora} and TenVOO~\cite{li2025parameter}.

\begin{table}[h]
\centering
\caption{Densities and coverages for different transfer learning strategies}
\begin{tabular}{lcc}
\toprule
\textbf{Frozen Layers} & \textbf{Density} & \textbf{Coverage} \\
\midrule
1st \& last residual blocks  & $0.599 \pm 0.081$ & $0.588 \pm 0.112$ \\
2nd \& the 2nd last blocks   & $0.596 \pm 0.088$ & $0.586 \pm 0.115$ \\
4th \& the 4th last blocks   & $0.600 \pm 0.081$ & $0.586 \pm 0.115$ \\
5th \& the 5th last blocks   & $0.599 \pm 0.087$ & $0.585 \pm 0.117$ \\
MLP layers                   & $0.597 \pm 0.083$ & $0.586 \pm 0.118$ \\
\bottomrule
\end{tabular}
\label{tab:density_coverage}
\end{table}


\section[More]{Pseudocode for Three Stages of GUST} 
\label{app:pseudocode}

To improve clarity and reproductivity, this section shows the pseudocode for the self-supervised pretraining (Alg.~\ref{alg:gust-pretrain}), fine-tuning (Alg.~\ref{alg:gust-finetune}), and sampling (Alg.~\ref{alg:gust-sampling}).

\begin{algorithm}[h]
\caption{Self-supervised pretraining}
\label{alg:gust-pretrain}
\begin{algorithmic}[1]
\Require Pretraining dataset $D_{\text{pretrain}}$; number of diffusion timesteps $T$; noise schedule $\{\beta_t\}_{t=1}^T$; U-Net with parameters $\theta$; number of training iterations $N_{\text{pretrain}}$; batch size $B$
\Ensure Pretrained U-Net weights $\theta_{\text{pre}}$
\State Define $\alpha_t \gets 1-\beta_t$ and $\bar{\alpha}_t \gets \prod_{s=1}^t \alpha_s$ for $t=1,\dots,T$
\State Initialize $\theta$
\For{$\text{iter}=1$ to $N_{\text{pretrain}}$}
  \State Sample minibatch $\mathcal{B}=\{(\mathbf{x}_{\text{nom}},\mathbf{x}_{\text{fab}})_i\}_{i=1}^B \subset D_{\text{pretrain}}$
  \ForAll{$(\mathbf{x}_{\text{nom}},\mathbf{x}_{\text{fab}}) \in \mathcal{B}$}
    \State Sample $t \sim \mathrm{Unif}\{1,\dots,T\}$, $\epsilon \sim \mathcal{N}(0,I)$
    \State $\mathbf{x}_t \gets \sqrt{\bar{\alpha}_t}\,\mathbf{x}_{\text{fab}} + \sqrt{1-\bar{\alpha}_t}\,\epsilon$
    \State $\hat{\epsilon} \gets \varepsilon_\theta\!\big(\mathbf{x}_t,\, t \, \big| \, \mathbf{x}_{\text{nom}}\big)$
    \State Update $\theta$ by Adam step on $\nabla_\theta \lVert \epsilon - \hat{\epsilon} \rVert^2$
  \EndFor
\EndFor
\State \Return $\theta_{\text{pre}} \gets \theta$
\end{algorithmic}
\end{algorithm}

\begin{algorithm}[h]
\caption{Fine-tuning}
\label{alg:gust-finetune}
\begin{algorithmic}[1]
\Require Pretrained weights $\theta_{\text{pre}}$; fine-tuning set $D_{\text{finetune}}$; number of diffusion timesteps $T$; noise schedule $\{\beta_t\}_{t=1}^T$; freezing strategy $\mathcal{F}$ (set of layers/blocks to freeze); number of training iterations $N_{\text{finetune}}$; batch size $B$
\Ensure Fine-tuned weights $\theta^\ast$
\State Define $\alpha_t \gets 1-\beta_t$ and $\bar{\alpha}_t \gets \prod_{s=1}^t \alpha_s$
\State $\theta \gets \theta_{\text{pre}}$; freeze parameters indicated by $\mathcal{F}$
\For{$\text{iter}=1$ to $N_{\text{finetune}}$}
  \State Sample minibatch $\mathcal{B}=\{(\mathbf{x}_{\text{nom}},\mathbf{x}_{\text{fab}})_j\}_{j=1}^B \subset D_{\text{finetune}}$
  \ForAll{$(\mathbf{x}_{\text{nom}},\mathbf{x}_{\text{fab}}) \in \mathcal{B}$}
    \State Sample $t \sim \mathrm{Unif}\{1,\dots,T\}$, $\epsilon \sim \mathcal{N}(0,I)$
    \State $\mathbf{x}_t \gets \sqrt{\bar{\alpha}_t}\,\mathbf{x}_{\text{fab}} + \sqrt{1-\bar{\alpha}_t}\,\epsilon$
    \State $\hat{\epsilon} \gets \varepsilon_\theta\!\big(\mathbf{x}_t,\, t \, \big| \, \mathbf{x}_{\text{nom}}\big)$
    \State Update \textbf{only unfrozen} parameters of $\theta$ by Adam step on $\nabla_\theta \lVert \epsilon - \hat{\epsilon} \rVert^2$
  \EndFor
\EndFor
\State \Return $\theta^\ast \gets \theta$
\end{algorithmic}
\end{algorithm}

\begin{algorithm}[h]
\caption{Sampling}
\label{alg:gust-sampling}
\begin{algorithmic}[1]
\Require Fine-tuned weights $\theta^\ast$; target nominal design $\mathbf{x}_{\text{nom}}$; number of diffusion timesteps $T$; noise schedule $\{\beta_t\}_{t=1}^T$; variance $\sigma_t^2$; number of generated samples $M$
\Ensure Generated samples $\{\hat{\mathbf{x}}^{(i)}\}_{i=1}^M$
\State Define $\alpha_t \gets 1-\beta_t$ and $\bar{\alpha}_t \gets \prod_{s=1}^t \alpha_s$
\For{$i = 1$ to $M$}
  \State Sample $\mathbf{x}_T \sim \mathcal{N}(\mathbf{0},\mathbf{I})$
  \For{$t = T, \dots, 1$}
    \State Sample $\mathbf{z} \sim \mathcal{N}(\mathbf{0},\mathbf{I})$ if $t>1$, else $\mathbf{z}=\mathbf{0}$
    \State $\mathbf{x}_{t-1} \gets \frac{1}{\sqrt{\alpha_t}}\left( \mathbf{x}_t - \frac{1-\alpha_t}{\sqrt{1-\bar{\alpha}_t}}\epsilon_{\theta^\ast}(\mathbf{x}_t, t \big| \mathbf{x}_\text{nom}) \right) + \sigma_t \mathbf{z}$
  \EndFor
  \State $\hat{\mathbf{x}}^{(i)} \gets \mathbf{x}_0$
\EndFor
\State \Return $\{\hat{\mathbf{x}}^{(i)}\}_{i=1}^M$
\end{algorithmic}
\end{algorithm}




\section[More]{Comparison to Augmented Direct Training} 
\label{app:augDT}

Data augmentation is another common approach to address data insufficiency. While augmenting the manufacturing data by applying random geometric transformation on the real geometries may synthetically create sufficient data to train the DDPM, the synthetic data being added will pollute the distribution of real-world as-fabricated geometries and therefore is against the purpose of learning real-world geometric uncertainty.

To show this, we added an \textit{augmented direct training} baseline (\emph{Augmented DT}) that trains the conditional DDPM from scratch on the \textit{augmented} real-world fine-tuning data. To augment the real as-fabricated geometry data, we apply the same random geometric transformations used to synthesize the pretraining data (Sec.~\ref{sec:datapreparation}). This process results in 62 perturbed geometries for each as-fabricated geometry. With 15 nominal designs in the fine-tuning data and the original 64 as-fabricated geometries per nominal design, we obtain an augmented dataset with $15\times 64\times (1+62)=60,480$ ``as-fabricated'' geometries in total. For a fair comparison, we ensure that this number is similar to the number of ``as-fabricated'' geometries in the pretraining data. The distributions of homogenized elastic properties (Fig.~\ref{fig: AugScratch}) show that \emph{Augmented DT} produces distributions closer to the ground truth than direct training (DT), possibly due to higher training stability and lower chances of overfitting. However, distributions of augmented DT exhibit much lower variances than the ground truth (similar to the distributions of pretraining data), indicating the risk of using hardcoded geometric variability to approximate the ground truth. On the other hand, GUST still aligns most closely to the ground truth compared to either DT or augmented DT.

\begin{figure*}[h]
\centering\includegraphics[width=1\linewidth]{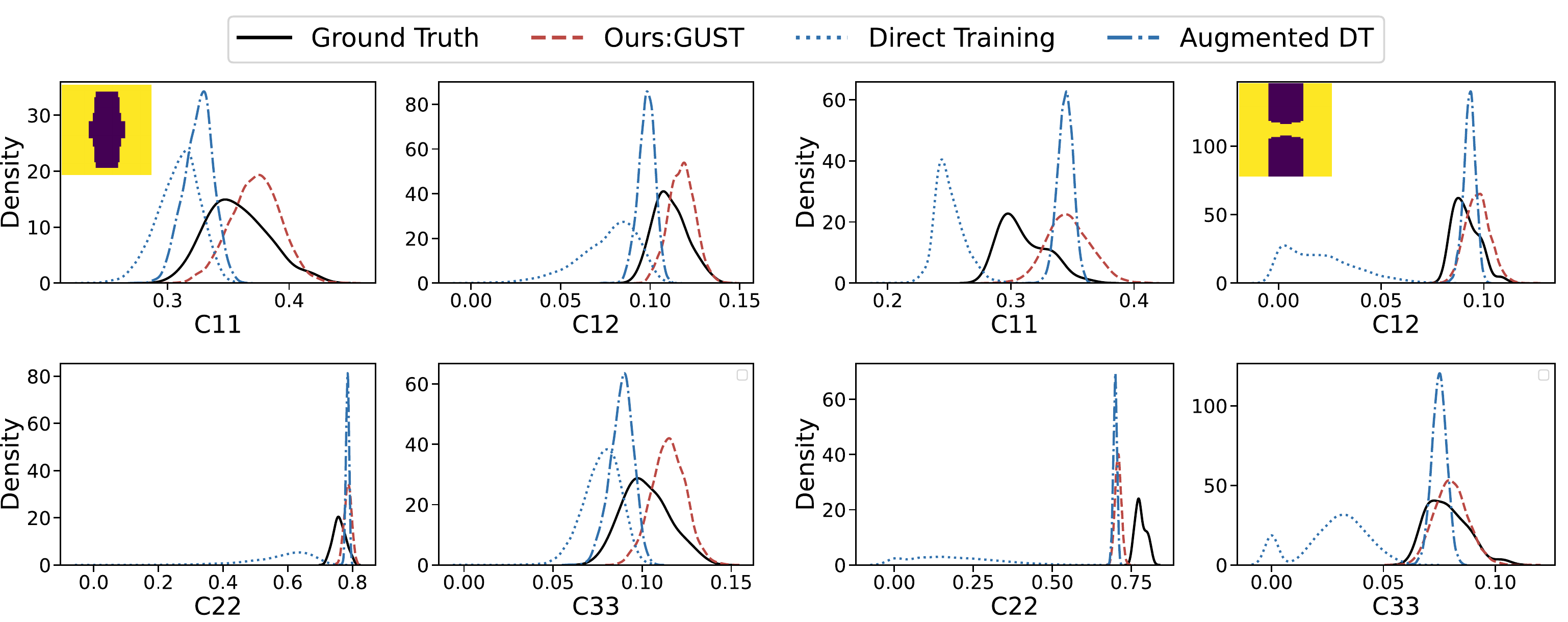}
\caption{KDEs of homogenized elastic tensor components ($C_{11}$, $C_{12}$, $C_{22}$, $C_{33}$) for two test unit cell designs in the real-world experiment. The corresponding nominal design is shown at the corner of each panel.}
\label{fig: AugScratch}
\end{figure*}

\end{document}